\documentclass{article}

\usepackage{iclr2027_conference,times}

\usepackage[utf8]{inputenc}
\usepackage[T1]{fontenc}
\usepackage[hidelinks]{hyperref}
\usepackage{url}
\usepackage{booktabs}
\usepackage{amsfonts}
\usepackage{enumitem}
\usepackage{nicefrac}
\usepackage{microtype}
\usepackage{xcolor}
\usepackage{graphicx}
\usepackage{multirow}
\usepackage{float}
\usepackage{amsmath}
\usepackage{amssymb}
\usepackage{mathtools}
\usepackage{amsthm}
\usepackage{dsfont}
\usepackage{xspace}
\usepackage{algorithm}
\usepackage{algpseudocode}
\usepackage{subcaption}
\usepackage{placeins}

\newtheorem{theorem}{Theorem}[section]  
\newtheorem{corollary}[theorem]{Corollary} 
\theoremstyle{definition}
\newtheorem{definition}{Definition}
\renewcommand{\headrulewidth}{0pt}

\newcommand{\methodname}{\textsc{Spelunka}\xspace}

\title{\methodname: A Framework for Finding Unanticipated Associations in Vision-Language Models}

\author{%
  Cassandra Parent \\
  MIT \\
  \texttt{cparent@mit.edu} \\
  \And
  Isha Agarwal \\
  MIT \\
  \texttt{agarwali@mit.edu} \\
  \And
  Kumail Alhamoud \\
  MIT \\ 
  \texttt{kumail@mit.edu} \\
  \And
  Charikleia Moraitaki \\
  MIT \\
  \texttt{hmor@mit.edu} \\
  \And 
  Rafiya Javed \\
  Google \\
  \texttt{rafiyajaved@google.com} \\
  \And 
  Walter Gerych* \\ 
  WPI \\
  \texttt{wgerych@wpi.edu} \\ 
  \And
  Marzyeh Ghassemi* \\
  MIT \\
  \texttt{mghassem@mit.edu} \\ \\
  *Co-corresponding
}

\begin{document}

\maketitle

\begin{abstract}
Vision-Language Models (VLMs) are the backbone of many state-of-the-art multimodal systems, but often encode spurious and harmful associations from their training data. While prior work focuses on removing known biases from VLMs, methods for discovering unanticipated associations for retrieval tasks remain underexplored. We introduce \methodname, an automated, inference-time framework for finding unanticipated distributional biases for retrieval tasks in VLMs without pre-specified labeled subgroups, inference-time human supervision, or model fine-tuning. 
Given a concept of interest, \methodname generates captions for related images, extracts relevant entities from captions, and selects semantically meaningful associations. A frequency-based detection rule then identifies spurious associations by measuring over- or under-representation of these attributes in a VLM's top-ranked outputs relative to their full dataset distribution. We evaluate \methodname across four datasets, and find that it 1) successfully recovers both synthetic and naturally occurring spurious correlations, 2) achieves an average AUROC of 0.93 across four datasets in identifying images strongly associated with a query, and 3) outperforms baselines in finding features that are strongly associated with a given query.
\end{abstract}
    
 \section{Introduction}
\label{sec:intro}

Vision-Language Models (VLMs) have quickly become the state-of-the-art for multimodal tasks \citep{radford2021learning, li2022blip, liu2024visual}, such as image retrieval \citep{lahajal2024enhancing}, zero and few-shot classification \citep{radford2021learning, an2023more}, and open-vocabulary object detection \citep{du2022learning}. However, VLMs can encode both spurious and harmful associations present in training data, leading to undesired behavior such as co-occurring object bias \citep{varma2024ravl}, failure to recognize objects in atypical settings \citep{arias2023assessing}, and perpetuating negative social stereotypes \citep{alabdulmohsin2024clip, kim2024discovering, friedrich2023fair, hamidieh2024identifying}. Given that VLMs are the backbones for such a wide range of multi-modal applications, it is crucial to remove these undesired associations, otherwise known as spurious correlations \citep{ye2024spuriouscorrelationsmachinelearning}. 

Prior work has largely focused on \emph{eliminating} known biases \citep{alabdulmohsin2024clip, kim2024discovering, berg-etal-2022-prompt, chuang2023debiasing, kong2024mitigating, wang2021gender, luo2024fairclip, zhu2023debiased, shen2023finetuning, jung2024unified, gerych2024bendvlm, dehdashtian2023fairvlm}, with relatively less targeting \emph{finding} unanticipated correlations in VLMs \citep{varma2024ravl, kruspe2024towards}. Many techniques assume prior knowledge of which attributes or concepts are spurious, in which case bias can be removed via fine-tuning \citep{zhu2023debiased, alabdulmohsin2024clip, shen2023finetuning} or post-hoc intervention \citep{jung2024unified, gerych2026wring, chuang2023debiasing, dehdashtian2023fairvlm}. However, prior work is limited to clearly defined tasks, e.g., classification, where differences in accuracy can be measured, but fails to explore tasks with less-defined metrics, such as retrieval \citep{eyuboglu2022domino, jain2022distilling, kim2024discovering, varma2024ravl}.

\begin{figure}[t]
  \begin{subfigure}[t]{\linewidth}
    \centering
    \includegraphics[width=0.95\linewidth]{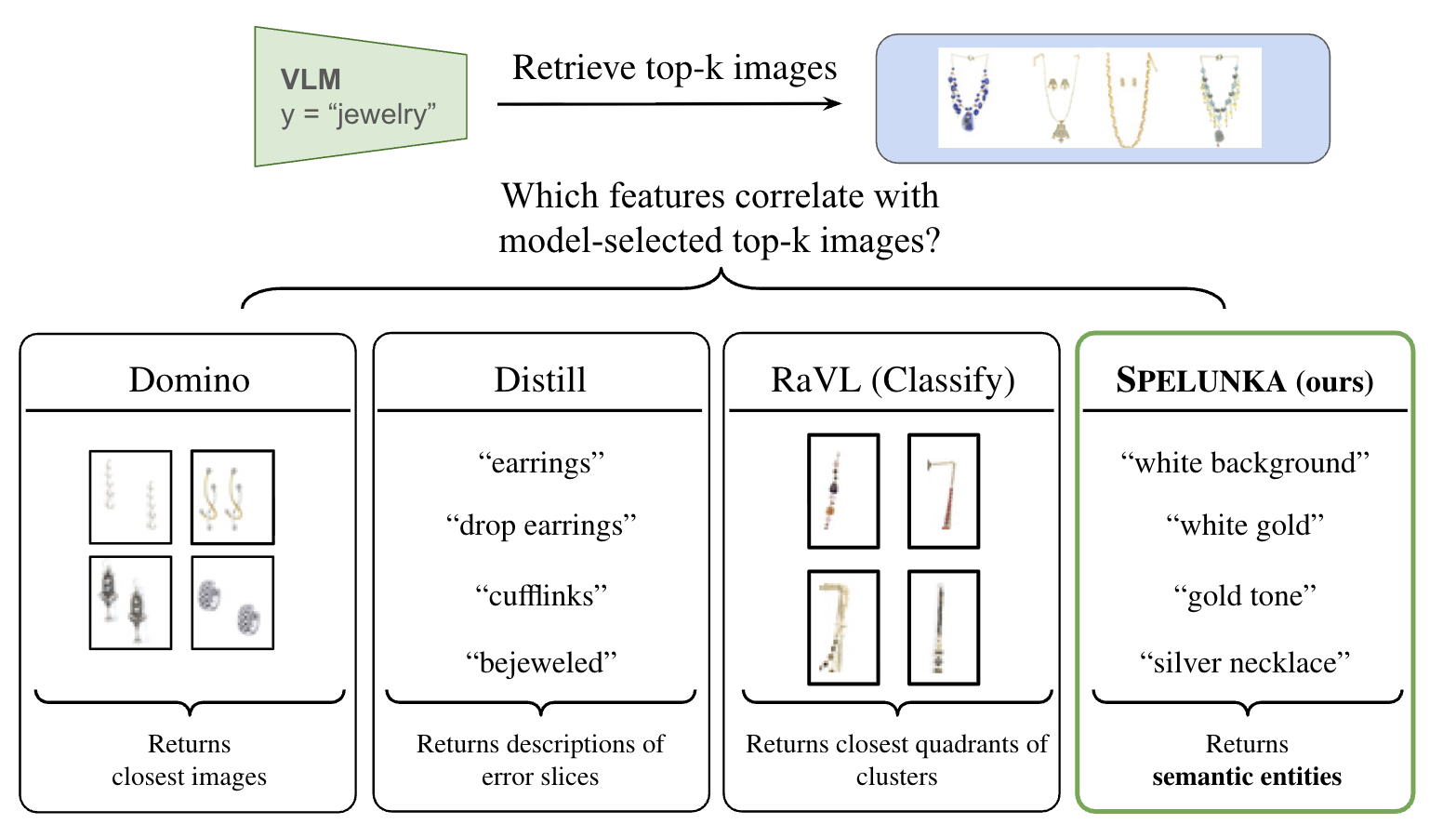}
        
  \caption{
  Identified biases 
  on "jewelry" concept with \texttt{clip-vit-base-patch32}. Compared to other methods, \methodname is able to pull out meaningful entities that map to the top retrieved images.}
  \label{fig:comparator}
  \end{subfigure}

  \caption{\textbf{\textsc{Spelunka} discovers and enables mitigation of unanticipated biases in VLM retrieval.} \textsc{Spelunka} differentiates itself by being able to work on tasks that (a) do not have labels and (b) are semantically meaningful for downstream users compared to other methods.}
  \label{fig:fig1}
\end{figure}

Our work focuses on finding \emph{unanticipated associations} in learned models, with three criteria. First, we do not assume access to existing labels or a classification task to generate spurious features or subgroups. Second, we avoid fine-tuning or auxiliary model training as this is computationally infeasible in many settings. Third, we seek a solution that provides semantically-meaningful identified features that can be easily assessed and removed by downstream debiasing techniques.

To address these criteria, we propose \methodname (\textbf{S}emantic \textbf{P}attern \textbf{E}xtraction for \textbf{L}ocating \textbf{Unk}nown \textbf{A}ssociations). To identify spurious correlations for a \emph{concept}, such as ``person'', \methodname compares the top-$k$ distribution of images related to the concept (e.g. top-$k$ images of people) to an overall distribution for that concept (e.g. a large dataset of people). Specifically, \methodname compares the relative frequency of different \emph{elements} of the images, such as co-occurring objects, expressions, and backgrounds. If an element is over- or under-represented in the top-$k$ retrieved images compared to its frequency within the full set of images for the target concept, then we can identify that the VLM is biased towards or against images where this element is present. To avoid the prohibitively expensive task of hand-labeling the elements of all images, \methodname uses a captioning model, and \methodname then extracts element labels and determines their significance. Importantly, we prove that \methodname\ does \emph{not} require its caption-generating component to be unbiased; the difference in concept frequency between the distributions can still identify spurious associations even with a biased image-captioning model. 

\textbf{Contributions.} 
The primary contributions of this work are:
\begin{itemize}[nosep]
    \item We propose \methodname, an automated spurious-correlation detection pipeline that can provide semantically-meaningful biases in retrieval tasks with no dataset labels.
    
    \item We demonstrate \methodname's ability to accurately identify biases in both synthetic spurious correlations (e.g., waterbirds and injected grayscale bias) \citep{rojas2022the, Sagawa2020Distributionally}, and naturally occurring dataset biases \citep{yang2023emoset, rojas2022the, param_aggarwal_2019, karkkainen2019fairface}.
    
    \item We validate that entities discovered by \methodname can identify whether an image is strongly associated with a query on a range of embedding models \citep{radford2021learning,fini2024multimodalautoregressivepretraininglarge,zhai2023sigmoidlosslanguageimage} and datasets \citep{yang2023emoset, rojas2022the, param_aggarwal_2019, karkkainen2019fairface}. On \texttt{clip-vit-base-patch-16}, We find that entities identified across four datasets and queries that describe the image type, i.e., ``jewelry'' in an image set of jewelry are predictive of top-$k$ images that will be associated with the term (average AUROC of 0.93 on 16 entities). On this task, \methodname outperforms four baselines: (\texttt{Cluster}, adapted from Domino \citep{eyuboglu2022domino}, \texttt{Distill} \citep{jain2022distilling}), and retrieval and classification versions of \texttt{RaVL} \citep{varma2024ravl}. 
    
\end{itemize}
 
\section{Related Work}
\label{sec:related}

\paragraph{Spurious Correlations in VLMs}
Spurious correlations, features that are coincidentally or confounded to be correlated with data labels, have been known to cause misleading or incorrect outcomes in machine learning models \citep{ye2024spuriouscorrelationsmachinelearning}. VLMs are a powerful model type that has grown in popularity due to state-of-the-art performance in both retrieval \citep{lahajal2024enhancing} and zero- and few-shot classification tasks \citep{radford2021learning, an2023more}. However, VLMs are still prone to spurious correlations which can lead to unsafe downstream performance, such as generating harmful content when faced with unsafe prompts \citep{chen2025safetymiragespuriouscorrelations}. Other undesired behavior from associations present in training data include co-occurring object bias (e.g. butterflies being associated with flowers \citep{varma2024ravl}), failure to recognize objects in atypical settings (e.g. removing backgrounds) \citep{arias2023assessing}, and perpetuating negative social stereotypes, (e.g. occupational gender bias) \citep{alabdulmohsin2024clip, kim2024discovering, friedrich2023fair}. While advances have been made in debiasing VLM embeddings for target concepts \citep{gerych2026wring, chuang2023debiasing}, determining the spurious correlations that VLM encodes is an ongoing challenge, as we do not have \emph{a priori} knowledge of all of the biases that the model encodes.

\paragraph{Finding Spurious Correlations}
Prior work in spurious correlation detection focuses on studying subsets of data where a model performs poorly on a defined classification task \citep{eyuboglu2022domino, jain2022distilling, kim2024discovering, varma2024ravl}. For instance, Domino \citep{eyuboglu2022domino} finds the images that perform poorly and then uses that class' proximity to cross-modal embeddings to describe that poor performing slice of data. While these approaches improve performance in the defined classification tasks, there remains a gap in finding spurious correlations in datasets where a clear performance assessment does not exist. \citet{dammu2023detectingspuriouscorrelationsrobust} developed a method that generates saliency maps to identify spurious correlations in an unlabeled set of images, but does not map these features to meaningful labels that can be understood and used for downstream tasks. This lack of semantically-meaningful features remains a challenge across all spurious correlation detection approaches where post-hoc explanations are unable to account for discovered artifacts that are unknown at test-time or are non-visible \citep{adebayo2022posthocexplanationsineffective}. \methodname aims to both identify spurious correlations with meaningful labels without specifying a classification task.

\section{Methodology}

\subsection{Preliminaries}
Let $M_\theta$ denote a pretrained Vision-Language Model (VLM), which takes an image $x \in \mathcal{X}$ and a text query $y \in \mathcal{Y}$ and returns a relevance score $M_\theta(x, y) \in \mathbb{R}$. Let $\mathcal{D} = \{x_i\}_{i=1}^n$ be a set of images associated with a query $y$ (e.g., \texttt{TV}), from which we aim to identify unanticipated, spuriously correlated visual or textual attributes. We define some additional terms specific to our problem set-up:

\begin{definition}[Caption Vocabulary]
    The \textit{Caption Vocabulary} $C$ is a set of all words that are used by the captioner to describe a set of images. Within this vocabulary, there exist a set of entities $\{e_1, e_2,\ldots e_m\}$. 
\end{definition}

\begin{definition}[Entity Indicator]
    We define $\ell_e(x) \in \{0, 1\}$ to be the (unobserved) indicator of whether image $x \in \mathcal{D}$ contains entity $e$ or not, i.e. $\ell(x) := \mathds{1}\{e \in x\}.$
\end{definition}

We assume no knowledge of any labels or \textit{C}. Specifically, we assume no access to $\ell_e(x)$.  Our goal then is to detect entities $e \in C$ that are over- or under-represented in the top-$k$ images ranked by $M_\theta$ for query $y$, relative to the full dataset. We determine over and under-representation using \textit{frequency bias}. 

\begin{definition}[Frequency Bias]
    \textit{Frequency bias} is the shift of entity frequencies between $\mathcal{D}$ and $\mathcal{D}_{\text{top-}k}$, the top-$k$ images according to $M_\theta(\cdot, y)$. 
\end{definition}

\subsection{\methodname: Uncovering Unanticipated Associations in VLMs}
\methodname consists of two stages: first we extract and group entities for a set of images. Then we identify any spurious correlations the VLM has between a target query $y$ and the identified entities. 

\subsubsection{Stage 1: Generating Pseudolabels}
\label{sec:pseudolabels}
Let $f_{\text{cap}}$ be a multimodal LLM image captioning model. We first apply $f_{\text{cap}}$ to each image $x \in \mathcal{D}$, generating a natural language caption $c_x = f_{\text{cap}}(x)$. The captions from $f_{\text{cap}}$ describe objects, attributes, and contextual elements in an open-vocabulary fashion.

Next, we extract entity mentions from each caption $c_x$ using an instruction-tuned language model $f_{\text{ent}}$. We prompt the model to identify discrete entities such as co-occurring objects (e.g., ``smartphone''), descriptors (e.g., ``flat-screen''), or settings (e.g., ``brick wall''). Our
entity-extraction prompt is given 
in the Supplement. This produces a candidate set of entity mentions $E_x = \{e_{1_x}, \ldots, e_{m_x}\}$ per image. 

To organize these entities, we use agglomerative clustering to collapse entities of identical meanings, i.e. ``television'', ``tv'', and ``TVs'' are equivalent. For each image $x$ and entity $e \in C$, we define a binary pseudolabel $\hat{\ell}_e(x) \in \{0,1\}$ indicating the presence of $e$ in the caption $c_x$. The pseudolabel function is defined as  $\hat{\ell}_e(x) := \mathds{1}\{e \in E_x\}.$

\begin{figure}[t]
  \centering
  \includegraphics[width=0.95\linewidth]{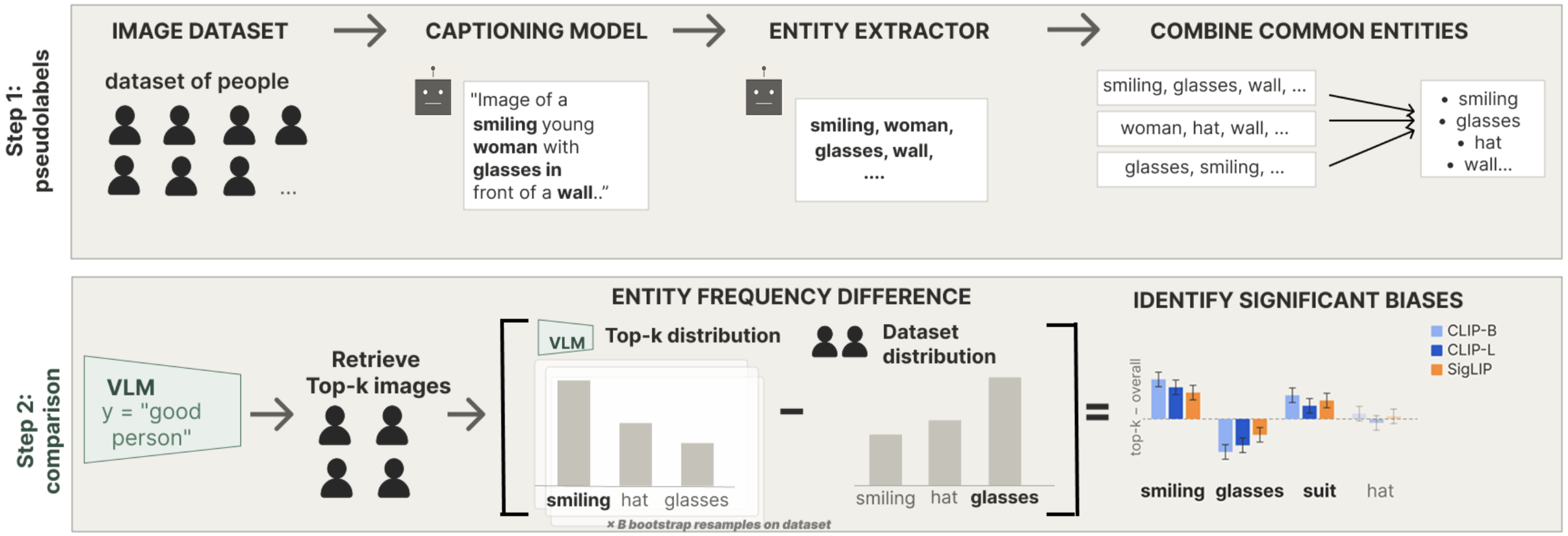}
\caption{\textbf{Overview of the \methodname\ pipeline.} Step 1 generates pseudolabels by captioning each image, extracting entities, and combines common entities into a final set. Step 2 retrieves the top-$k$ images for a query using the VLM and identifies spurious correlations as the difference between the entity distribution in the top-$k$ set and the full dataset distribution.}  \label{fig:overview}
\end{figure}

This step results in a weakly supervised binary matrix $L \in \{0,1\}^{|\mathcal{D}| \times |E|}$, where $L_{ij} = \hat{\ell}_{e_j}(x_i)$ captures the presence of entity $e_j$ in image $x_i$. This scalable and automatic extraction of labels from VLM-generated captions does not rely on predefined subgroups or human annotation (Figure \ref{fig:overview}).

\subsubsection{Stage 2: Identifying Over-Representation}

Given a distribution of pseudolabels for the presence of entities across the image set $\mathcal{D}$ for query $y$, we now identify which 
entities are spuriously over-represented in the VLM's top-ranked outputs.

Let $M_\theta$ be the vision-language model under audit. Given the query $y$, we compute the VLM relevance score $M_\theta(x, y)$ for each image $x \in \mathcal{D}$. We then rank all images in descending order of their relevance score and select the top-$k$ images: $\mathcal{D}_{\text{top-}k} := \text{Top-}k\left(\{M_\theta(x, y) : x \in \mathcal{D}\}\right)$.

For each entity $e \in E_C$, we compare its empirical prevalence in $\mathcal{D}_{\text{top-}k}$ against its prevalence in the full dataset $\mathcal{D}$. These quantities are estimated from the pseudolabel matrix $L \in \{0,1\}^{n \times |E|}$, where $L_{ij} = \hat{\ell}_{e_j}(x_i)$.

Define:
\begin{align*}
p_{\text{obs}}^{\mathcal{D}}(e) := \frac{1}{|\mathcal{D}|} \sum_{x \in \mathcal{D}} \hat{\ell}_e(x), \quad
p_{\text{obs}}^{\mathcal{D}_{\text{top-}k}}(e) := \frac{1}{|\mathcal{D}_{\text{top-}k}|} \sum_{x \in \mathcal{D}_{\text{top-}k}} \hat{\ell}_e(x), \quad
\Delta_{\text{obs}}(e) := p_{\text{obs}}^{\mathcal{D}_{\text{top-}k}}(e) - p_{\text{obs}}^{\mathcal{D}}(e)
\end{align*}

The over-representation score $\Delta_{\text{obs}}(e)$ quantifies the degree to which $e$ appears more frequently in the top-$k$ ranked images than would be expected based on its global frequency, i.e., a positive $\Delta_{\text{obs}}(e)$ is positive spurious association between entity $e$ and query $y$ in the model $M_\theta$.

This comparison is performed independently for each entity $e$, allowing us to identify unanticipated correlations across categories without relying on any prior knowledge of what the ``spurious'' attributes might be. Further, to report entities that are relevant we perform $N=1000$ bootstraps with replacement on the dataset on the identified entities. In practice, we report the associations we identify as significant, i.e. 95\% confidence intervals generated from the bootstrap do not cross zero. All graphs are sorted by the entities with the largest magnitude in one of the tested models with some minimum threshold, 5\% shift unless stated otherwise. 

\subsection{Robustness to Captioning Model Bias}
\methodname's pseudolabels come from a captioner that may itself be biased. It may over-mention some entities and under-mention others. Does this captioner bias corrupt the spurious correlations \methodname reports? We show that it does not. As long as the captioner detects each entity better than chance, the sign of $\Delta_{\text{obs}}(e)$ matches the sign of $\Delta_{\text{true}}(e)$, the \emph{direction} of bias is preserved.

We demonstrate that an imperfect pseudolabeling process can identify frequency bias by assuming a stochastic pseudolabeler with entity-dependent true positive rate (TPR) and false positive rate (FPR):
\begin{align*}
    TPR_e &:= \mathbb{E}[\hat{\ell}_e(x) \mid \ell_e(x) = 1] := \alpha_e, \quad
    FPR_e :=  \mathbb{E}[\hat{\ell}_e(x) \mid \ell_e(x) = 0] := \beta_e 
\end{align*}

We express the expected value of the pseudolabel as:
\[
\mathbb{E}[\hat{\ell}_e(x)] = \alpha_e \cdot \ell_e(x) + \beta_e \cdot (1 - \ell_e(x)) = \beta_e + (\alpha_e - \beta_e) \cdot \ell_e(x).
\]

Note that this allows for the pseudolabeling process to be biased; the TPR and FPR may be different for different entities. Let 
the expected observed prevalence be:
\[
\mathbb{E}[p_{\text{obs}}^{\mathcal{D}}(e)] := \frac{1}{|\mathcal{D}|} \sum_{x \in \mathcal{D}} \mathbb{E}[\hat{\ell}_e(x)] = \beta_e + (\alpha_e - \beta_e) \cdot p_{\text{true}}^{\mathcal{D}}(e).
\]

We compare the prevalence of $e$ between the full dataset $\mathcal{\mathcal{D}}$ and a subset $\mathcal{D}_{\text{top-}k}$ of the $k$ most relevant images for some query $y$. Let:

\begin{align*}
    \Delta_{\text{true}}(e) &:= p_{\text{true}}^{\mathcal{D}_{\text{top-}k}}(e) - p_{\text{true}}^{\mathcal{D}}(e),  \quad
    \Delta_{\text{obs}}^{\mathbb{E}}(e) := \mathbb{E}[p_{\text{obs}}^{\mathcal{D}_{\text{top-}k}}(e) - p_{\text{obs}}^{\mathcal{D}}(e)]
\end{align*}

By linearity of expectation and the above noise model:
\begin{align*}
\Delta_{\text{obs}}^{\mathbb{E}}(e)
&= \left[ \beta_e + (\alpha_e - \beta_e) \cdot p_{\text{true}}^{\mathcal{D}_{\text{top-}k}}(e) \right] - \left[ \beta_e + (\alpha_e - \beta_e) \cdot p_{\text{true}}^{\mathcal{D}}(e) \right] 
= (\alpha_e - \beta_e) \cdot \Delta_{\text{true}}(e).
\end{align*}

Thus, as long as the pseudolabeler performs better than random --- that is, $\alpha_e > \beta_e$ --- the expected observed over-representation $\Delta_{\text{obs}}^{\mathbb{E}}(e)$ has the same sign as the true over-representation $\Delta_{\text{true}}(e)$. This implies that the \emph{direction} of bias is preserved under label noise. Even with imperfect heuristics, consistent over-representation in the observed labels (e.g., $\Delta_{\text{obs}}^{\mathbb{E}}(e) > 0$) implies true over-representation ($\Delta_{\text{true}}(e) > 0$). 
Further analyses on the mathematical guarantees 
are in the Appendix.

\section{Experimental Design}
\label{experiments}
\textbf{\methodname Pipeline Setup.} 
In our implementation of \textsc{Spelunka}, we selected BLIP~\citep{li2022blip} as the captioning model $f_{\text{cap}}$ and Llama 3.1 \citep{grattafiori2024llama} as the entity extraction and category-labeling model $f_{\text{ent}}$ as small, local models. For entity embeddings, all-MiniLM-L6 \citep{reimers-2019-sentence-bert} and SpaCy \citep{spacey} were used as inputs into Scikit-learn's \citep{pedregosa2011scikit} Agglomerative Clustering, which was used with a 20\% distance threshold. Entities must have appeared at least 10 times in each dataset. We note that while Theorem E.1 demonstrates that meaningful biases can be found even under noisy processes, we demonstrate that a variety of captioning models (\texttt{llava-1.5-7b-hf} \citep{liu2024improvedbaselinesvisualinstruction} and \texttt{Qwen3-VL-8B-Instruct} \citep{yang2025qwen3technicalreport}) yield similar results Figure \ref{fig:combined}. Additional details are available in the Appendix.

\textbf{Models.}
We test \methodname on a range of CLIP \citep{radford2021learning} (\texttt{clip-vit-base-patch16}, \texttt{clip-vit-base-patch32}, \texttt{CLIP-ViT-L-14-laion2B-s32B-b82K}) an AIMV2 model (\texttt{aimv2-large-patch14-224-lit}) \citep{fini2024multimodalautoregressivepretraininglarge}, and a SigLip model (\texttt{siglip-base-patch16-224}) \citep{zhai2023sigmoidlosslanguageimage}.

\textbf{Data.}
We evaluate models on the \texttt{Fashion Product} \citep{param_aggarwal_2019}, \texttt{FairFace}, \citep{karkkainen2019fairface}, \texttt{Dollar Street} \citep{rojas2022the}, and \texttt{EmoSet} \citep{yang2023emoset} each with a maximum of 128 output tokens.
We extract eight concepts across the four datasets to evaluate \methodname. Concepts included were jewelry from \texttt{Fashion Product}, TV, street view, and home from \texttt{Dollar Street}, person and face from \texttt{FairFace}, and scene and person from \texttt{EmoSet}. Concepts were selected based on being an identifiable object with sufficient sample size with additional dataset descriptions in the Supplement. Note that \texttt{Fashion Product} and \texttt{EmoSet} differ from the other datasets by including ground truth subclass labels (e.g. rings, bracelets) for each concept (e.g. jewelry) in \texttt{Fashion Product} and different emotions for \texttt{EmoSet}. 
\section{Experimental Results}

\subsection{Finding Contextual Spurious Correlations}
\begin{figure}[t]
    \centering
    \includegraphics[width=\linewidth]{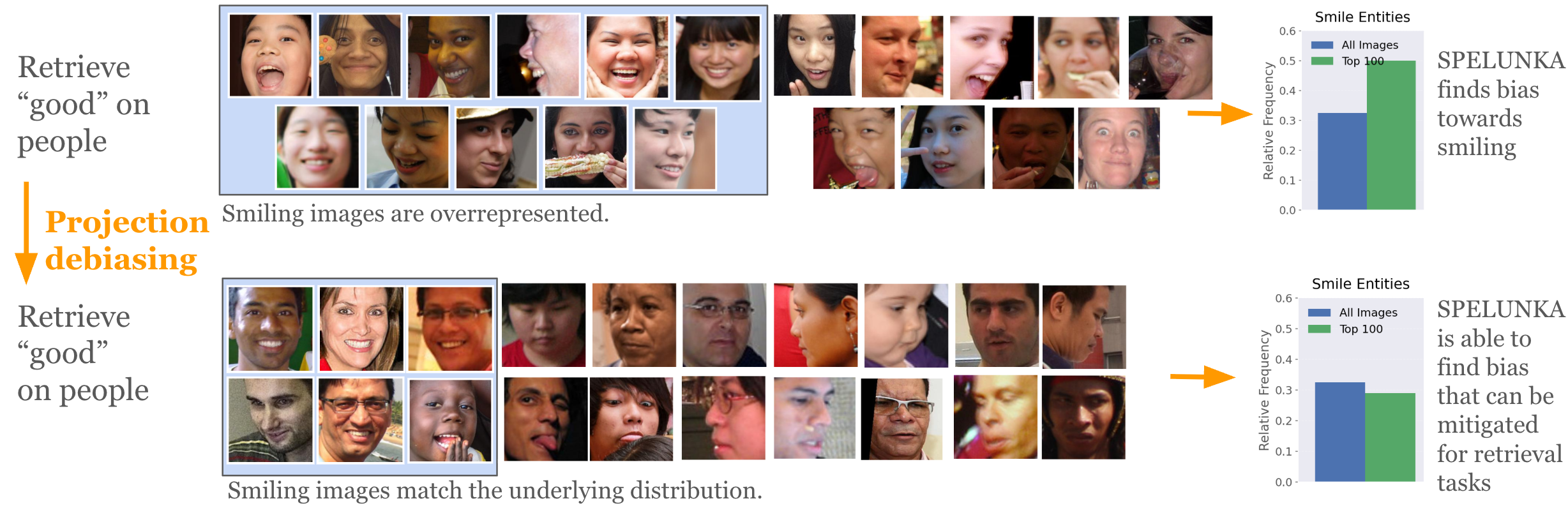}
    \caption{\textbf{\textsc{Spelunka} surfaces meaningful bias automatically by comparing entity frequencies between top-$k$ retrieved images and the full dataset, with no labels, task definition, or fine-tuning required.} 
    Querying AimV2 for ``good'' on \texttt{FairFace} overrepresents smiling faces relative to the dataset distribution (top). Projection debiasing, a common technique to remove a bias by orthogonalizing its embeddings to a concept \citep{bolukbasi2016man}, then aligns the retrieved distribution with the base rate (bottom).
    }
    \label{fig:gap}
   
\end{figure}

Unlike prior work, \methodname is able to detect spurious correlations without a defined classification task or labels. This enables bias detection in real-world subjective cases. For instance, we find that \methodname falsely attributes the entity ``smiling'' with ``good person'' in the \texttt{FairFace} dataset (Figure \ref{fig:gap}). Further, this bias, once identified, can be removed via projection debiasing \citep{bolukbasi2016man}. This demonstrates \methodname's ability to find actionable biases within models.  

\begin{figure}[t]
    \centering
    \includegraphics[width=\linewidth]{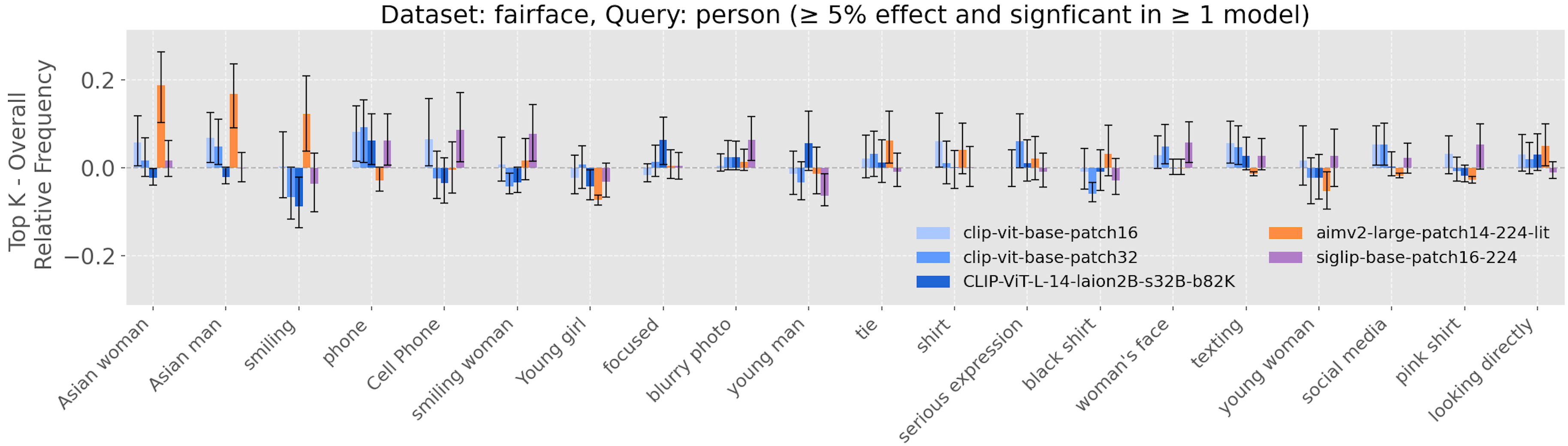}
    \caption{\textbf{Biases identified by \methodname on Fairface when querying ``person'' across models.} Images with entities \textit{Asian woman}, \textit{Asian man}, and \textit{smiling} appear more frequently than in the full dataset than for AIMV2.}
\label{fig:tv-person}
\end{figure}

Beyond a meaningful example, we demonstrate that the extracted labels expose over/under representations of elements in VLMs. On the \texttt{FairFace} dataset, we set the query $y$ to ``person.'' There were 20 entities that had a greater than 5\% over/under-representation and showed significance in at least one model (Figure \ref{fig:tv-person}). Asian women, Asian men, and smiling were all entities that were overrepresented in the \texttt{AIM-V2} model, meaning that if the image was an Asian man or woman, it was more likely to be retrieved by the VLM. Results for all datasets can be found in the Appendix. 

To show that the biases identified by \methodname are biases found within the model, we take the significant entities and apply them to a different dataset of people images, \texttt{UTKFace} \citep{zhifei2017cvpr}. We kept the same query and only looked at entities that appeared in both datasets. We found that 14/15 overlapping entities that were significant in \texttt{FairFace} had sign agreement with \texttt{UTKFace} (Table \ref{tab:utk}). This demonstrates that the associations found by \textsc{Spelunka} are biases of the \emph{model}, and that they can hold under varying applications.

\begin{table}[!ht]
    \centering
    \scriptsize
    \begin{tabular}{lllll}
    \hline
        \textbf{Entity} & \textbf{Model} & \textbf{FairFace}  & \textbf{UTKFace} & \textbf{Agreement} \\ \hline
        Asian woman & aimv2-large-patch14-224-lit & 0.187 & 0.168 & TRUE \\ 
        Asian man & aimv2-large-patch14-224-lit & 0.169 & 0.119 & TRUE \\ 
        smiling & aimv2-large-patch14-224-lit & 0.123 & 0.060 & TRUE \\ 
        smiling & CLIP-ViT-L-14-laion2B-s32B-b82K & -0.087 & 0.020 & FALSE \\ 
        smiling woman & siglip-base-patch16-224 & 0.077 & 0.150 & TRUE \\ 
        Young girl & aimv2-large-patch14-224-lit & -0.073 & -0.051 & TRUE \\ 
        Asian man & clip-vit-base-patch16 & 0.069 & 0.019 & TRUE \\ 
        young man & siglip-base-patch16-224 & -0.064 & -0.037 & TRUE \\ 
        tie & aimv2-large-patch14-224-lit & 0.062 & 0.063 & TRUE \\ 
        shirt & clip-vit-base-patch16 & 0.061 & 0.004 & TRUE \\ 
        serious expression & clip-vit-base-patch32 & 0.061 & 0.063 & TRUE \\ 
        black shirt & clip-vit-base-patch32 & -0.059 & -0.019 & TRUE \\ 
        Asian woman & clip-vit-base-patch16 & 0.057 & 0.048 & TRUE \\
        young woman & aimv2-large-patch14-224-lit & -0.053 & -0.024 & TRUE \\ 
        looking directly & aimv2-large-patch14-224-lit & 0.050 & 0.051 & TRUE \\ \hline
    \end{tabular}
    
    \caption{\textbf{\textsc{Spelunka} finds biases that hold across models.} From the overlapping entities in \texttt{FairFace} and \texttt{UTKFace}, that met at least a 5\% distribution shift and were significant in \texttt{FairFace}, 14/15 had sign agreement with \texttt{UTKFace}.}
    \label{tab:utk}
\end{table}

\subsection{Validation of Label Recovery and Spurious Correlation Detection}
\begin{figure}[t]
  \centering

  \begin{subfigure}[t]{0.48\linewidth}
    \centering
    \includegraphics[width=\linewidth]{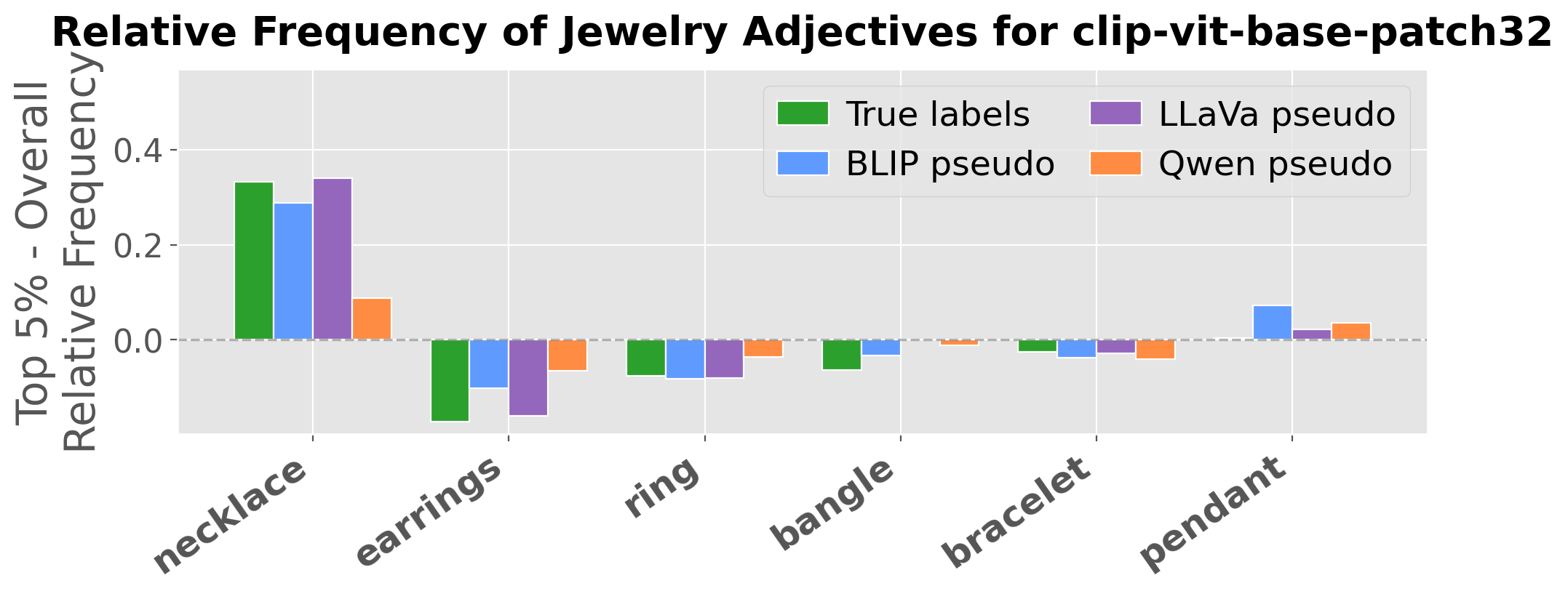}
\caption{Pseudolabels and true label frequency for subcategories in the Jewelry concept. \methodname accurately identifies true labels.}    \label{fig:psuedo-jewelery}
  \end{subfigure}
  \hfill
  \begin{subfigure}[t]{0.48\linewidth}
    \centering
    \includegraphics[width=\linewidth]{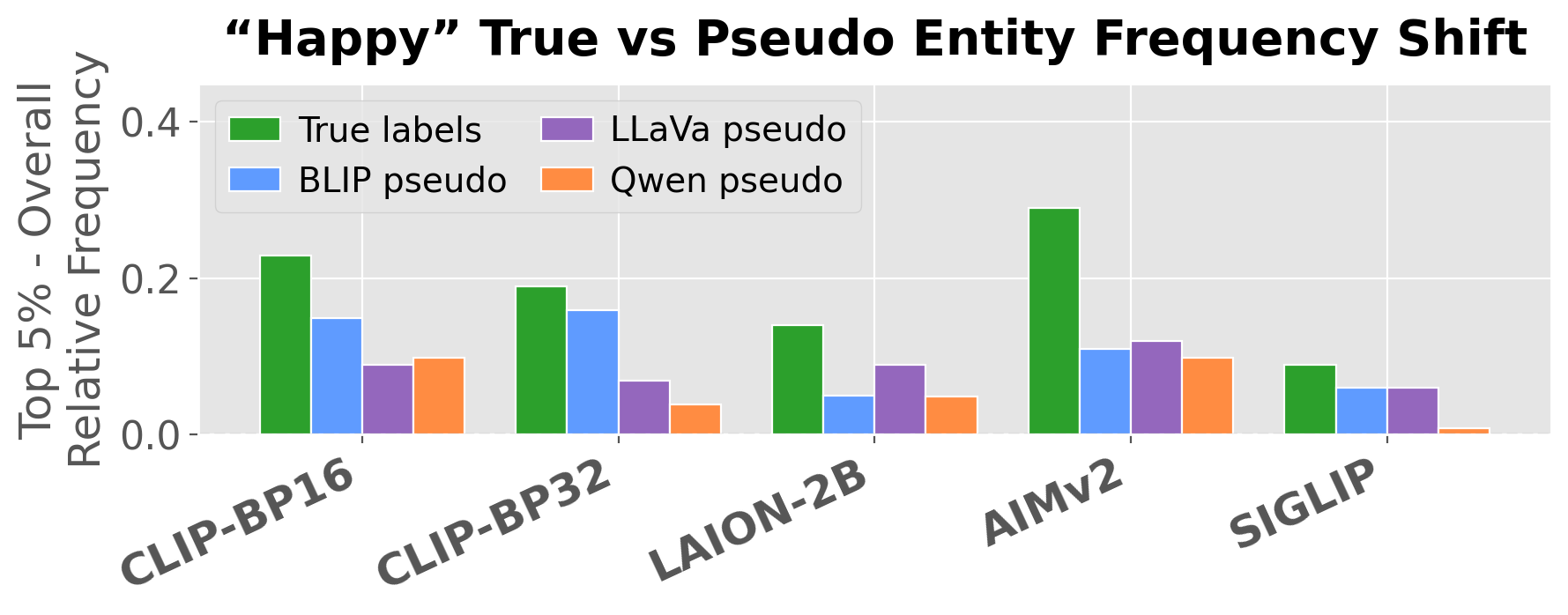}
\caption{Pseudolabels across models for \texttt{EmoSet} on `happy' emotion. \methodname identifies this subtle feature across models.}    \label{fig:pseudo-emo}
  \end{subfigure}  
\caption{\textbf{\methodname recovers ground-truth labels across datasets and captioning models.} Pseudolabel frequencies track true label frequencies on (a) \texttt{Fashion Product} jewelry subtypes and on (b) \texttt{EmoSet} `happy' images across CLIP and AIMv2 backbones.}  \label{fig:combined}
\end{figure}

\paragraph{Validation of Label Recovery.} \label{sec:step1_val}
We validate that \methodname correctly recovers ground-truth element labels. We use the jewelry images from the Fashion Product Images dataset, as this dataset contains labels for jewelry subtype (ring, necklace, etc). We also use the images from \texttt{EmoSet} labeled with a `happy' facial expression for validation on large images. As shown in Figure \ref{fig:combined}, \methodname's pseudolabels closely track ground-truth label frequencies, and this recovery is robust across captioning models (BLIP, Llava, Qwen) and VLM backbones (CLIP, AIMv2, SigLip) with models showing broad sign agreement. Results with additional datasets are in the Appendix.

\begin{figure}[t]
  \centering
  \includegraphics[width=0.85\linewidth]{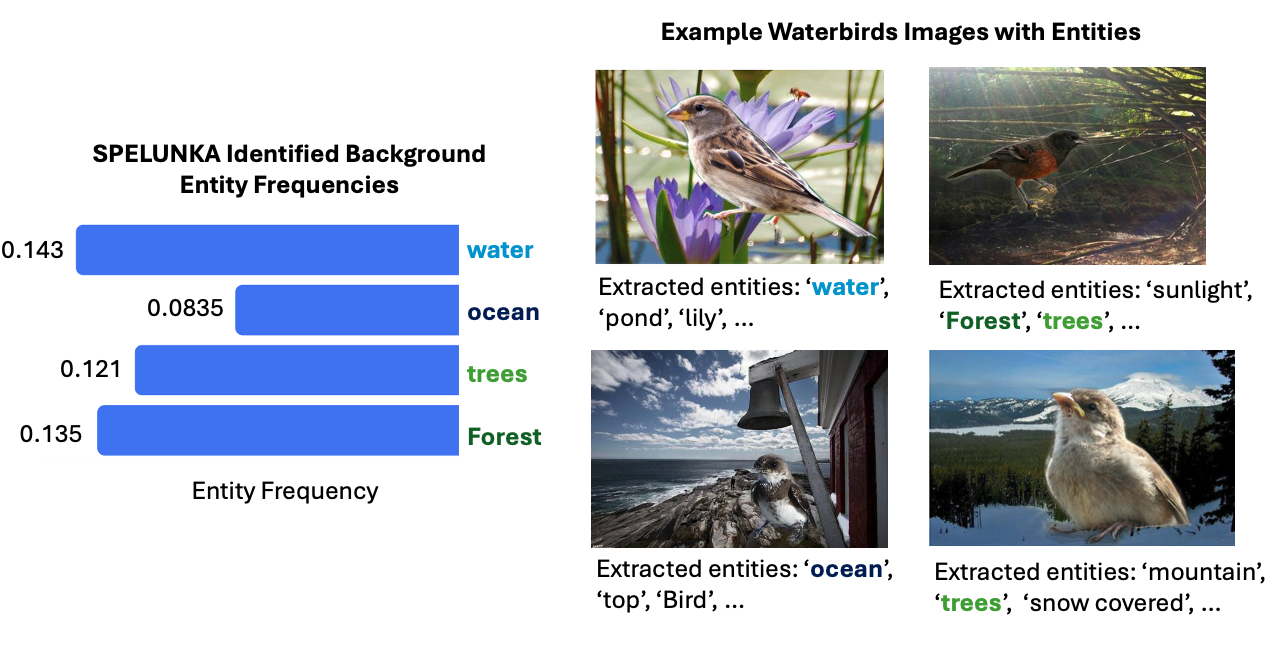}
   \caption{\textbf{\methodname recovers the known background spurious correlation in \texttt{Waterbirds}.} Without access to bird-type or background labels, \methodname extracts background entities (\textit{water}, \textit{ocean}, \textit{trees}, \textit{Forest}) at high frequency from the test set.}  \label{fig:waterbirds}
\end{figure}

\paragraph{Validation of Spurious Correlation Detection.} \label{sec:step2_val} 
We evaluated \methodname's ability to identify a known bias, by examining \texttt{Waterbirds} \citep{Sagawa2020Distributionally}. This dataset contains training data where bird type matches background in the training data and testing data where bird and background type are mixed. We find that \methodname extracts frequently occurring entities from the test dataset, such as `water', `ocean', `trees' and `Forest'. This shows \methodname's ability to identify known background spurious correlations with bird type (Figure \ref{fig:waterbirds}) and demonstrates that \methodname is able to uncover known spurious correlations from unlabeled datasets. We show an additional inserted bias experiment in the Appendix. 

\subsection{\textsc{Evaluating \textbf{Spelunka}} for Downstream Concept Feature Identification}
We compare \methodname against other spurious correlation identification methods. For each method, we train a classifier on the identified spurious features to predict whether a given input will appear in the top $5\%$ set of images. Higher AUROC indicates that the features identified by the method are more strongly correlated with the query.

As no other automated methods exist for unlabeled bias detection in VLMs for retrieval tasks, we construct four baselines from prior works. \texttt{Domino} is inspired by \citep{eyuboglu2022domino}. Here, we cluster image embeddings and then use the similarity to the \emph{mean} embedding of each cluster as a feature. We vary the number of clusters from 2 to 16. \texttt{Distill} \citep{jain2022distilling} is adapted by identifying spurious features through first identifying instances that are misclassified by a logistic regression classifier (e.g. misclassifications in jewelry types of rings, bracelets, necklaces, etc.). We then find natural language phrases from an LLM-generated corpus whose embeddings have the highest similarity to the normal vector. For each image embedding, we use similarity to these identified phrases as features, varying the number of phrases from 2 to 16. We use two versions of \texttt{RaVL} \citep{varma2024ravl} where images are segmented into quadrants prior to clustering. For \texttt{RaVL (retrieve)}, we select the number of clusters based on the number of tested features. 
\texttt{RaVL (classify)}
selects 256 features during clustering and then greedily selects the top features based on classifications of sublabels. Note that
\texttt{Distill} and \texttt{RaVL (classify)} only work on the \texttt{Fashion} and \texttt{EmoSet} dataset which has subtype labels to construct error slices from.

For \methodname, we define features as the similarity to the embeddings of the entities with the highest magnitude of difference between top-$k$ and overall prevalence, according to \methodname's pseudo-labeling process. We vary the number of entities from 2 to 16. We report AUROC \citep{Melo2013} on identifying top-$k$ images using a random forest \citep{randomforest} classifier and report confidence intervals using 500 bootstrapped samples over the test set (50/50 split due to sample size). Table \ref{tab:compare} quantitatively demonstrates \methodname's ability to identify spurious features that are strongly correlated with the query, with an average AUROC of 0.93 across the full set of queries using 16 entities (Table \ref{tab:compare-full}). We also find that the features are on average, more strongly correlated with the query than those identified by other methods. Further, \methodname's features are semantically meaningful. For instance, on \texttt{EmoSet}'s person concept, \methodname tells us top potentially spurious features are `smiling', `young girl', and `white shirt'. Notably, these are spurious semantic concepts. In contrast, other methods return numeric vectors that are traced back to regions of the original image for interpretation or error descriptions from a pre-defined classification task (Figure \ref{fig:comparator}). These results indicate that \methodname finds features that are both \emph{quantitatively and qualitatively} more \emph{informative} than the features identified by other methods. 
Additional results on the full set of queries are in the Appendix. 

\begin{table}[h]
\centering
\caption{\label{tab:compare} \textbf{\methodname outperforms baselines with an average AUROC of 0.93 in identifying images strongly associated with a query using a random forest classifier across $4$ datasets and $6$ queries with 16 entities} (full results in Appendix). Intervals denote $95\%$ confidence intervals with $500$ bootstrapped samples. 
Comparisons done with \texttt{clip-base-patch-16}. \texttt{Distill} and \texttt{RaVL (Classify)}  require subtype labels and thus cannot be applied to all datasets.} 
\scriptsize
\setlength{\tabcolsep}{6pt}
\begin{tabular}{@{}ccccccc@{}}
\toprule
                            & \# Entities & \texttt{Domino} & \texttt{RaVL} (Retrieve) & \texttt{Distill} & \texttt{RaVL} (Classify) & \methodname \\ \midrule
\multirow{4}{*}{\shortstack{Fashion,\\Jewelry}} 
                            & 2          & 0.61 [0.54, 0.68] & 0.55 [0.50, 0.58] & \textbf{0.63 [0.56, 0.71]} & 0.59 [0.54, 0.63] & 0.56 [0.48, 0.65] \\
                            & 4          & 0.68 [0.61, 0.75] & 0.67 [0.63, 0.71] & 0.67 [0.58, 0.75] & 0.66 [0.57, 0.66] & \textbf{0.70 [0.63, 0.77]}     \\
                            & 8          & 0.73 [0.65, 0.79] & 0.67 [0.63, 0.70] &  \textbf{0.78 [0.72, 0.84]}  & 0.72 [0.68, 0.75] &  0.75 [0.68, 0.81]    \\
                            & 16         & 0.73 [0.66, 0.80]  & 0.69 [0.65, 0.72] &  0.80 [0.74, 0.87]  & 0.74 [0.70, 0.77] & \textbf{0.93 [0.90, 0.97]} \\ \midrule
\multirow{4}{*}{\shortstack{EmoSet,\\Person}} 
                            & 2          & 0.77 [0.72, 0.81] & 0.50 [0.47, 0.53] &  0.76 [0.71, 0.81] & 0.60 [0.57, 0.63] &  \textbf{0.86 [0.81, 0.90]} \\
                            & 4          & 0.85 [0.82, 0.89] & 0.73 [0.71, 0.75] & 0.83 [0.79, 0.87] & 0.73 [0.71, 0.76] & \textbf{0.94 [0.91, 0.96] }    \\
                            & 8          & 0.864 [0.82, 0.90] & 0.73 [0.71, 0.76] & 0.87 [0.83, 0.91]  & 0.74 [0.71, 0.76] &  \textbf{0.95 [0.93, 0.97]}     \\
                            & 16         & 0.86 [0.82, 0.91]  & 0.75 [0.72, 0.77] & 0.88 [0.84, 0.92]  & 0.79 [0.77, 0.81] & \textbf{0.97 [0.95, 0.98]} \\ \midrule
\multirow{4}{*}{\shortstack{Dollar\\Street, TV}}
                            & 2          & 0.67 [0.55, 0.78] & 0.54 [0.48, 0.61] & - & - & \textbf{0.89 [0.80, 0.96]}     \\
                            & 4          & 0.53 [0.40, 0.64] & 0.50 [0.45, 0.57] &  -  & - & \textbf{0.87 [0.76, 0.95]}  \\
                            & 8          & 0.74 [0.62, 0.84] & 0.59 [0.53, 0.65] &  - & - & \textbf{0.92 [0.86, 0.96]}     \\
                            & 16         & 0.82 [0.73, 0.90] & 0.61 [0.55, 0.67] & - & - & \textbf{0.92 [0.84, 0.96]}     \\ \midrule
\multirow{4}{*}{\shortstack{FairFace,\\Face}} 
                            & 2          & 0.63 [0.58, 0.69] & 0.51 [0.48, 0.54] & - & - & \textbf{0.83 [0.78, 0.87]}     \\
                            & 4          & 0.71 [0.65, 0.76] & 0.57 [0.55, 0.60] & -  & - & \textbf{0.88 [0.84, 0.91]}  \\
                            & 8          &  0.73 [0.67, 0.78] & 0.66 [0.63, 0.69] &  - & - & \textbf{0.90 [0.87, 0.94]} \\
                            & 16         & 0.72 [0.67, 0.77] & 0.67 [0.64, 0.70] &  - & - & \textbf{0.97 [0.96, 0.98]}  \\ \midrule
\end{tabular}
\end{table}

\section{Limitations and Future Work}
\label{sec:limitations}
Although \methodname provides a practical way to surface unanticipated associations in VLMs, it has limitations. First, while we demonstrate that directional bias of captioning models is preserved under label noise and empirically demonstrate that varying captioning models yield similar pseudo-label results, the entities extracted will have variation dependent on caption quality. Future work could experiment with taking the super set of several captioning model entities to stabilize this.
Second, \methodname finds correlations based on frequency but does not assess their harmfulness or contextual relevance.  It is a valuable auditing tool, but is not a substitute for human judgment in high-stakes applications. Despite these limitations, \methodname has the potential for meaningful impact due to its automated ability to identify spurious correlations regardless of the task. As multimodal models see wider real-world deployment, accessible auditing tools become increasingly important. \methodname lowers the barrier for auditing pretrained multimodal models, and we hope this work sparks further research on systematic, open set methods for identifying and addressing 
unwanted associations.

\section{Conclusions}
We introduced \methodname, a fully automated framework for uncovering unanticipated associations in vision-language models (VLMs). \methodname identifies spurious correlations without labeled subgroups, fine-tuning, or manual inspection. Our method leverages open-vocabulary captioning and large language models to construct semantically meaningful spurious correlations. Through theoretical analysis and empirical validation, we demonstrated that \methodname can reliably identify meaningful biases across multiple domains, often outperforming baselines in identifying features predictive of being in the retrieval set. By enabling scalable and unsupervised auditing of VLMs, \methodname represents a step towards ensuring the robustness, generalizability, and fairness of multimodal AI systems in real-world applications.

\subsubsection*{Acknowledgments}
This work was supported in part by a National Science Foundation CAREER Award (2339381), and an AI2050 Award Early Career Fellowship (G-25-68042), a Sloan Research Fellow Award (FG-2025-24277), a MIT-Google Computing Innovation Award, and a NVIDIA Academic Grants Program award. This material is based upon work supported by the National Science Foundation Graduate Research Fellowship under Grant No. (2141064).

{
\small
\bibliographystyle{iclr2027_conference}
\bibliography{main}

@String(CVPR= {IEEE Conf. Comput. Vis. Pattern Recog.})

@String(AAAI = {AAAI})

@String(CVPR  = {CVPR})

@article{hamidieh2024identifying, title={Identifying Implicit Social Biases in Vision-Language Models}, volume={7}, url={https://ojs.aaai.org/index.php/AIES/article/view/31657}, DOI={10.1609/aies.v7i1.31657}, number={1}, journal={Proceedings of the AAAI/ACM Conference on AI, Ethics, and Society}, author={Hamidieh, Kimia and Zhang, Haoran and Gerych, Walter and Hartvigsen, Thomas and Ghassemi, Marzyeh}, year={2024}, month={Oct.}, pages={547-561} }

@article{chuang2023debiasing,
  title={Debiasing vision-language models via biased prompts},
  author={Chuang, Ching-Yao and Jampani, Varun and Li, Yuanzhen and Torralba, Antonio and Jegelka, Stefanie},
  journal={arXiv preprint arXiv:2302.00070},
  year={2023}
}

@inproceedings{li2022blip,
  title={Blip: Bootstrapping language-image pre-training for unified vision-language understanding and generation},
  author={Li, Junnan and Li, Dongxu and Xiong, Caiming and Hoi, Steven},
  booktitle={International conference on machine learning},
  pages={12888--12900},
  year={2022},
  organization={PMLR}
}

@article{liu2024visual,
  title={Visual instruction tuning},
  author={Liu, Haotian and Li, Chunyuan and Wu, Qingyang and Lee, Yong Jae},
  journal={Advances in neural information processing systems},
  volume={36},
  year={2024}
}

@misc{yang2025qwen3technicalreport,
      title={Qwen3 Technical Report}, 
      author={An Yang and Anfeng Li and Baosong Yang and Beichen Zhang and Binyuan Hui and Bo Zheng and Bowen Yu and Chang Gao and Chengen Huang and Chenxu Lv and Chujie Zheng and Dayiheng Liu and Fan Zhou and Fei Huang and Feng Hu and Hao Ge and Haoran Wei and Huan Lin and Jialong Tang and Jian Yang and Jianhong Tu and Jianwei Zhang and Jianxin Yang and Jiaxi Yang and Jing Zhou and Jingren Zhou and Junyang Lin and Kai Dang and Keqin Bao and Kexin Yang and Le Yu and Lianghao Deng and Mei Li and Mingfeng Xue and Mingze Li and Pei Zhang and Peng Wang and Qin Zhu and Rui Men and Ruize Gao and Shixuan Liu and Shuang Luo and Tianhao Li and Tianyi Tang and Wenbiao Yin and Xingzhang Ren and Xinyu Wang and Xinyu Zhang and Xuancheng Ren and Yang Fan and Yang Su and Yichang Zhang and Yinger Zhang and Yu Wan and Yuqiong Liu and Zekun Wang and Zeyu Cui and Zhenru Zhang and Zhipeng Zhou and Zihan Qiu},
      year={2025},
      eprint={2505.09388},
      archivePrefix={arXiv},
      primaryClass={cs.CL},
      url={https://arxiv.org/abs/2505.09388}, 
}

@inproceedings{
  Sagawa2020Distributionally,
  title={Distributionally Robust Neural Networks},
  author={Shiori Sagawa and Pang Wei Koh and Tatsunori B. Hashimoto and Percy Liang},
  booktitle={International Conference on Learning Representations},
  year={2020},
  url={https://openreview.net/forum?id=ryxGuJrFvS}
}

@misc{zhifei2017cvpr,
  title={Age Progression/Regression by Conditional Adversarial Autoencoder},
  author={Zhifei Zhang and Yang Song and Hairong Qi},
  booktitle={IEEE Conference on Computer Vision and Pattern Recognition (CVPR)},
  year={2017},
  organization={IEEE}
}

@misc{liu2024improvedbaselinesvisualinstruction,
      title={Improved Baselines with Visual Instruction Tuning}, 
      author={Haotian Liu and Chunyuan Li and Yuheng Li and Yong Jae Lee},
      year={2024},
      eprint={2310.03744},
      archivePrefix={arXiv},
      primaryClass={cs.CV},
      url={https://arxiv.org/abs/2310.03744}, 
}

@article{kong2024mitigating,
  title={Mitigating test-time bias for fair image retrieval},
  author={Kong, Fanjie and Yuan, Shuai and Hao, Weituo and Henao, Ricardo},
  journal={Advances in Neural Information Processing Systems},
  volume={36},
  year={2024}
}

@misc{kim2024discovering,
      title={Discovering and Mitigating Visual Biases through Keyword Explanation}, 
      author={Younghyun Kim and Sangwoo Mo and Minkyu Kim and Kyungmin Lee and Jaeho Lee and Jinwoo Shin},
      year={2024},
      eprint={2301.11104},
      archivePrefix={arXiv},
      primaryClass={cs.LG}
}

@inproceedings{berg-etal-2022-prompt,
    title = "A Prompt Array Keeps the Bias Away: Debiasing Vision-Language Models with Adversarial Learning",
    author = "Berg, Hugo  and
      Hall, Siobhan  and
      Bhalgat, Yash  and
      Kirk, Hannah  and
      Shtedritski, Aleksandar  and
      Bain, Max",
    editor = "He, Yulan  and
      Ji, Heng  and
      Li, Sujian  and
      Liu, Yang  and
      Chang, Chua-Hui",
    booktitle = "Proceedings of the 2nd Conference of the Asia-Pacific Chapter of the Association for Computational Linguistics and the 12th International Joint Conference on Natural Language Processing (Volume 1: Long Papers)",
    month = nov,
    year = "2022",
    address = "Online only",
    publisher = "Association for Computational Linguistics",
    url = "https://aclanthology.org/2022.aacl-main.61",
    pages = "806--822",
}

@inproceedings{radford2021learning,
  title={Learning transferable visual models from natural language supervision},
  author={Radford, Alec and Kim, Jong Wook and Hallacy, Chris and Ramesh, Aditya and Goh, Gabriel and Agarwal, Sandhini and Sastry, Girish and Askell, Amanda and Mishkin, Pamela and Clark, Jack and others},
  booktitle={International conference on machine learning},
  pages={8748--8763},
  year={2021},
  organization={PMLR}
}

@article{karkkainen2019fairface,
  title={Fairface: Face attribute dataset for balanced race, gender, and age},
  author={K{\"a}rkk{\"a}inen, Kimmo and Joo, Jungseock},
  journal={arXiv preprint arXiv:1908.04913},
  year={2019}
}

@article{lahajal2024enhancing,
  title={Enhancing Image Retrieval: A Comprehensive Study on Photo Search using the CLIP Mode},
  author={Lahajal, Naresh Kumar and others},
  journal={arXiv preprint arXiv:2401.13613},
  year={2024}
}

@article{an2023more,
  title={More context, less distraction: Visual classification by inferring and conditioning on contextual attributes},
  author={An, Bang and Zhu, Sicheng and Panaitescu-Liess, Michael-Andrei and Mummadi, Chaithanya Kumar and Huang, Furong},
  journal={arXiv preprint arXiv:2308.01313},
  year={2023}
}

@inproceedings{dehdashtian2023fairvlm,
  title={FairVLM: Mitigating Bias In Pre-Trained Vision-Language Models},
  author={Dehdashtian, Sepehr and Wang, Lan and Boddeti, Vishnu},
  booktitle={The Twelfth International Conference on Learning Representations},
  year={2023}
}

@inproceedings{zhu2023debiased,
  title={Debiased fine-tuning for vision-language models by prompt regularization},
  author={Zhu, Beier and Niu, Yulei and Lee, Saeil and Hur, Minhoe and Zhang, Hanwang},
  booktitle={Proceedings of the AAAI Conference on Artificial Intelligence},
  year={2023}
}

@article{luo2024fairclip,
  title={FairCLIP: Harnessing Fairness in Vision-Language Learning},
  author={Luo, Yan and Shi, Min and Khan, Muhammad Osama and Afzal, Muhammad Muneeb and Huang, Hao and Yuan, Shuaihang and Tian, Yu and Song, Luo and Kouhana, Ava and Elze, Tobias and others},
  journal={arXiv preprint arXiv:2403.19949},
  year={2024}
}

@article{alabdulmohsin2024clip,
  title={CLIP the Bias: How Useful is Balancing Data in Multimodal Learning?},
  author={Alabdulmohsin, Ibrahim and Wang, Xiao and Steiner, Andreas and Goyal, Priya and D'Amour, Alexander and Zhai, Xiaohua},
  journal={arXiv preprint arXiv:2403.04547},
  year={2024}
}

@article{shen2023finetuning,
  title={Finetuning Text-to-Image Diffusion Models for Fairness},
  author={Shen, Xudong and Du, Chao and Pang, Tianyu and Lin, Min and Wong, Yongkang and Kankanhalli, Mohan},
  journal={arXiv preprint arXiv:2311.07604},
  year={2023}
}

@article{bolukbasi2016man,
  title={Man is to computer programmer as woman is to homemaker? debiasing word embeddings},
  author={Bolukbasi, Tolga and Chang, Kai-Wei and Zou, James Y and Saligrama, Venkatesh and Kalai, Adam T},
  journal={Advances in neural information processing systems},
  volume={29},
  year={2016}
}

@article{wang2021gender,
  title={Are gender-neutral queries really gender-neutral? mitigating gender bias in image search},
  author={Wang, Jialu and Liu, Yang and Wang, Xin Eric},
  journal={arXiv preprint arXiv:2109.05433},
  year={2021}
}

@article{friedrich2023fair,
  title={Fair diffusion: Instructing text-to-image generation models on fairness},
  author={Friedrich, Felix and Brack, Manuel and Struppek, Lukas and Hintersdorf, Dominik and Schramowski, Patrick and Luccioni, Sasha and Kersting, Kristian},
  journal={arXiv preprint arXiv:2302.10893},
  year={2023}
}

@inproceedings{gerych2024bendvlm,
  title={BendVLM: Test-Time Debiasing of Vision-Language Embeddings},
  author={Gerych, Walter and Zhang, Haoran and Hamidieh, Kimia and Pan, Eileen and Sharma, Maanas and Hartvigsen, Thomas and Ghassemi, Marzyeh},
  booktitle={The Thirty-eighth Annual Conference on Neural Information Processing Systems},
  year={2024}
}

@inproceedings{jung2024unified,
  title={A Unified Debiasing Approach for Vision-Language Models across Modalities and Tasks},
  author={Jung, Hoin and Jang, Taeuk and Wang, Xiaoqian},
  booktitle={The Thirty-eighth Annual Conference on Neural Information Processing Systems},
  year={2024}
}

@inproceedings{du2022learning,
  title={Learning to prompt for open-vocabulary object detection with vision-language model},
  author={Du, Yu and Wei, Fangyun and Zhang, Zihe and Shi, Miaojing and Gao, Yue and Li, Guoqi},
  booktitle={Proceedings of the IEEE/CVF conference on computer vision and pattern recognition},
  pages={14084--14093},
  year={2022}
}

@article{varma2024ravl,
  title={Ravl: Discovering and mitigating spurious correlations in fine-tuned vision-language models},
  author={Varma, Maya and Delbrouck, Jean-Benoit and Chen, Zhihong and Chaudhari, Akshay and Langlotz, Curtis},
  journal={Advances in Neural Information Processing Systems},
  volume={37},
  pages={82235--82264},
  year={2024}
}

@article{arias2023assessing,
  title={Assessing biases through visual contexts},
  author={Arias-Duart, Anna and Gimenez-Abalos, Victor and Cort{\'e}s, Ulises and Garcia-Gasulla, Dario},
  journal={Electronics},
  volume={12},
  number={14},
  pages={3066},
  year={2023},
  publisher={MDPI}
}

@article{kruspe2024towards,
  title={Towards detecting unanticipated bias in Large Language Models},
  author={Kruspe, Anna},
  journal={arXiv preprint arXiv:2404.02650},
  year={2024}
}

@article{grattafiori2024llama,
  title={The llama 3 herd of models},
  author={Grattafiori, Aaron and Dubey, Abhimanyu and Jauhri, Abhinav and Pandey, Abhinav and Kadian, Abhishek and Al-Dahle, Ahmad and Letman, Aiesha and Mathur, Akhil and Schelten, Alan and Vaughan, Alex and others},
  journal={arXiv preprint arXiv:2407.21783},
  year={2024}
}

@article{jain2022distilling,
  title={Distilling model failures as directions in latent space},
  author={Jain, Saachi and Lawrence, Hannah and Moitra, Ankur and Madry, Aleksander},
  journal={arXiv preprint arXiv:2206.14754},
  year={2022}
}

@article{eyuboglu2022domino,
  title={Domino: Discovering systematic errors with cross-modal embeddings},
  author={Eyuboglu, Sabri and Varma, Maya and Saab, Khaled and Delbrouck, Jean-Benoit and Lee-Messer, Christopher and Dunnmon, Jared and Zou, James and R{\'e}, Christopher},
  journal={arXiv preprint arXiv:2203.14960},
  year={2022}
}

@misc{ye2024spuriouscorrelationsmachinelearning,
      title={Spurious Correlations in Machine Learning: A Survey}, 
      author={Wenqian Ye and Guangtao Zheng and Xu Cao and Yunsheng Ma and Aidong Zhang},
      year={2024},
      eprint={2402.12715},
      archivePrefix={arXiv},
      primaryClass={cs.LG},
      url={https://arxiv.org/abs/2402.12715}, 
}

@misc{chen2025safetymiragespuriouscorrelations,
      title={Safety Mirage: How Spurious Correlations Undermine VLM Safety Fine-tuning}, 
      author={Yiwei Chen and Yuguang Yao and Yihua Zhang and Bingquan Shen and Gaowen Liu and Sijia Liu},
      year={2025},
      eprint={2503.11832},
      archivePrefix={arXiv},
      primaryClass={cs.AI},
      url={https://arxiv.org/abs/2503.11832}, 
}

@misc{adebayo2022posthocexplanationsineffective,
      title={Post hoc Explanations may be Ineffective for Detecting Unknown Spurious Correlation}, 
      author={Julius Adebayo and Michael Muelly and Hal Abelson and Been Kim},
      year={2022},
      eprint={2212.04629},
      archivePrefix={arXiv},
      primaryClass={cs.LG},
      url={https://arxiv.org/abs/2212.04629}, 
}

@misc{dammu2023detectingspuriouscorrelationsrobust,
      title={Detecting Spurious Correlations via Robust Visual Concepts in Real and AI-Generated Image Classification}, 
      author={Preetam Prabhu Srikar Dammu and Chirag Shah},
      year={2023},
      eprint={2311.01655},
      archivePrefix={arXiv},
      primaryClass={cs.LG},
      url={https://arxiv.org/abs/2311.01655}, 
}

@inproceedings{
rojas2022the,
title={The Dollar Street Dataset: Images Representing the Geographic and Socioeconomic Diversity of the World},
author={William A Gaviria Rojas and Sudnya Diamos and Keertan Ranjan Kini and David Kanter and Vijay Janapa Reddi and Cody Coleman},
booktitle={Thirty-sixth Conference on Neural Information Processing Systems Datasets and Benchmarks Track},
year={2022},
url={https://openreview.net/forum?id=qnfYsave0U4}
}

@misc{zhai2023sigmoidlosslanguageimage,
      title={Sigmoid Loss for Language Image Pre-Training}, 
      author={Xiaohua Zhai and Basil Mustafa and Alexander Kolesnikov and Lucas Beyer},
      year={2023},
      eprint={2303.15343},
      archivePrefix={arXiv},
      primaryClass={cs.CV},
      url={https://arxiv.org/abs/2303.15343}, 
}

@inproceedings{reimers-2019-sentence-bert,
  title = "Sentence-BERT: Sentence Embeddings using Siamese BERT-Networks",
  author = "Reimers, Nils and Gurevych, Iryna",
  booktitle = "Proceedings of the 2019 Conference on Empirical Methods in Natural Language Processing",
  month = "11",
  year = "2019",
  publisher = "Association for Computational Linguistics",
  url = "https://arxiv.org/abs/1908.10084",
}

@article{spacey,
    author = {Honnibal, Matthew and Montani, Ines and Van Landeghem, Sofie and Boyd, Adriane},
    title = "spaCy: Industrial-strength Natural Language Processing in Python",
    year = 2020,
    doi = "10.5281/zenodo.1212303"
}

@article{pedregosa2011scikit,
  title={Scikit-learn: Machine learning in Python},
  author={Pedregosa, Fabian and Varoquaux, Ga{\"e}l and Gramfort, Alexandre and Michel, Vincent and Thirion, Bertrand and Grisel, Olivier and Blondel, Mathieu and Prettenhofer, Peter and Weiss, Ron and Dubourg, Vincent and others},
  journal={the Journal of machine Learning research},
  volume={12},
  pages={2825--2830},
  year={2011},
  publisher={JMLR. org}
}

@misc{param_aggarwal_2019,
	title={Fashion Product Images Dataset},
	url={https://www.kaggle.com/ds/139630},
	DOI={10.34740/KAGGLE/DS/139630},
	publisher={Kaggle},
	author={Param Aggarwal},
	year={2019}
}

@misc{fini2024multimodalautoregressivepretraininglarge,
      title={Multimodal Autoregressive Pre-training of Large Vision Encoders}, 
      author={Enrico Fini and Mustafa Shukor and Xiujun Li and Philipp Dufter and Michal Klein and David Haldimann and Sai Aitharaju and Victor Guilherme Turrisi da Costa and Louis Béthune and Zhe Gan and Alexander T Toshev and Marcin Eichner and Moin Nabi and Yinfei Yang and Joshua M. Susskind and Alaaeldin El-Nouby},
      year={2024},
      eprint={2411.14402},
      archivePrefix={arXiv},
      primaryClass={cs.CV},
      url={https://arxiv.org/abs/2411.14402}, 
}

@Inbook{Melo2013,
author="Melo, Francisco",
editor="Dubitzky, Werner
and Wolkenhauer, Olaf
and Cho, Kwang-Hyun
and Yokota, Hiroki",
title="Area under the ROC Curve",
bookTitle="Encyclopedia of Systems Biology",
year="2013",
publisher="Springer New York",
address="New York, NY",
pages="38--39",
isbn="978-1-4419-9863-7",
doi="10.1007/978-1-4419-9863-7_209",
url="https://doi.org/10.1007/978-1-4419-9863-7_209"
}

@misc{randomforest,
author="Breiman, Leo",
title="Random Forests",
year="2001",
publisher="Springer Nature Link",
doi="https://doi.org/10.1023/A:1010933404324"
}

@inproceedings{yang2023emoset,
title={EmoSet: A Large-scale Visual Emotion Dataset with Rich Attributes},
author={Yang, Jingyuan and Huang, Qirui and Ding, Tingting and Lischinski, Dani and Cohen-Or, Danny and Huang, Hui},
booktitle={Proceedings of the IEEE/CVF International Conference on Computer Vision},
pages={20383--20394},
year={2023}
}

@inproceedings{gerych2026wring,
  title={WRING OUT THE BIAS: A ROTATION-BASED ALTER-NATIVE TO PROJECTION DEBIASING},
  author={Gerych, Walter and Parent, Cassandra and Quinn Perian, Rafiya Javed and Solomon, Justin and Ghassemi, Marzyeh},
  booktitle={The Fourteenth International Conference on Learning Representations},
  year={2026}
}
}

\clearpage
\appendix

\section{Experiment Set-Up Additional Details}
\label{sec:exp-set-up}
While we select specific models in our pipeline setup in Section \ref{experiments}, we note that \methodname can work using a variety of captioning and extraction methods. We picked models that were relatively small and local for ease of reproducibility and low computational cost. 

The following prompt was used for entity extraction \ref{sec:pseudolabels}: ``I am going to give you a caption of a picture of \{concept\}. Please extract all of all objects and descriptors from the caption and return them as a comma separated list. Each list item should be a MAXIMUM of two words. Respond ONLY with this list. Here is the description for you to work with: ''

\section{Dataset Additional Details}
\label{sec:data-details}
\begin{figure}[h]
    \includegraphics[width=1.0\linewidth]{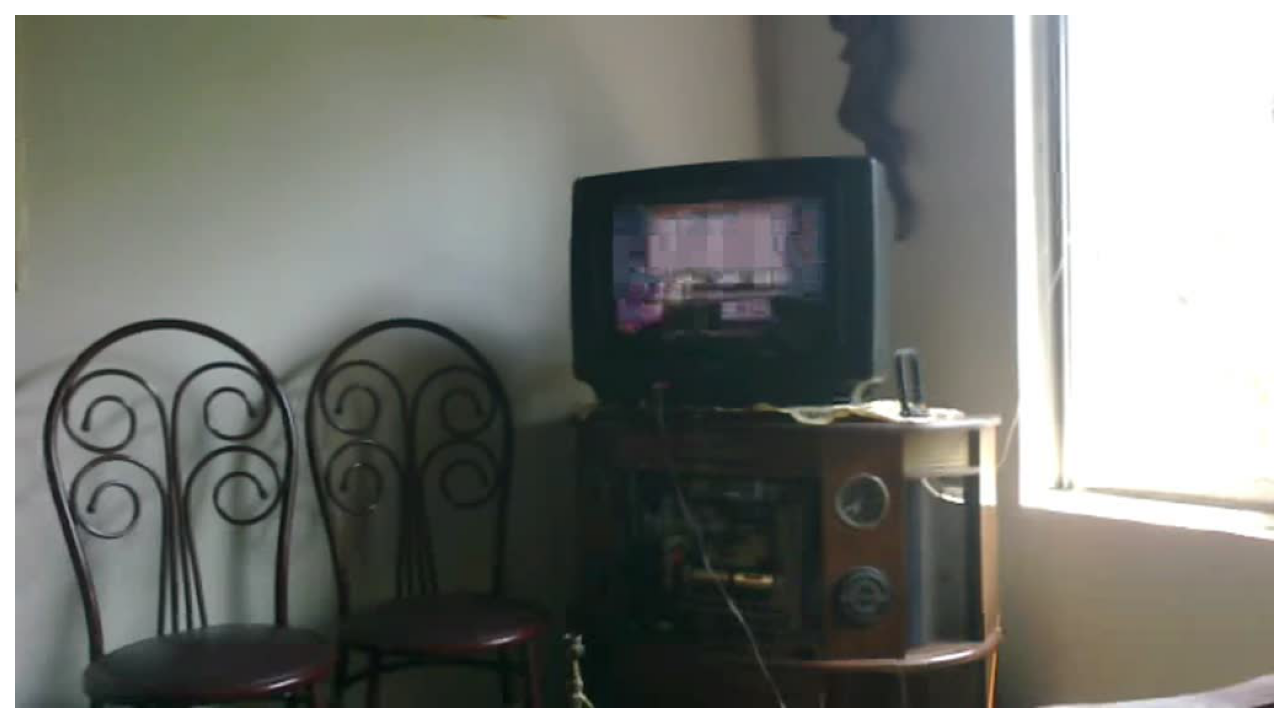}
    \centering
    \caption{Example image from `TV' concept in \texttt{Dollar Street}. }
    \label{fig:tv_example}
\end{figure}

\begin{figure}[h]
    \includegraphics[width=1.0\linewidth]{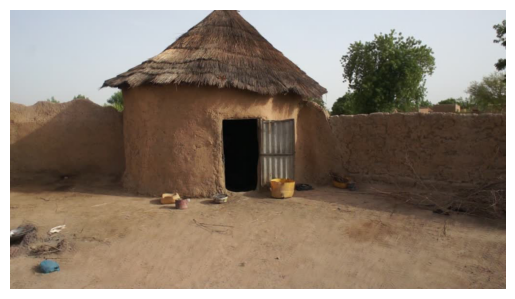}
    \centering
    \caption{Example image from `home' concept in \texttt{Dollar Street}. }
    \label{fig:home_example}
\end{figure}

\begin{figure}[h]
    \includegraphics[width=0.5\linewidth]{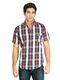}
    \centering
    \caption{Example image from `topwear' concept in \texttt{Fashion Product}. }
    \label{fig:topwear_example}
\end{figure}

\begin{figure}[h]
    \includegraphics[width=0.5\linewidth]{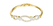}
    \centering
    \caption{Example image from `jewelry' concept in \texttt{Fashion Product}. }
    \label{fig:jewelry_example}
\end{figure}

\begin{figure}[h]
    \includegraphics[width=0.5\linewidth]{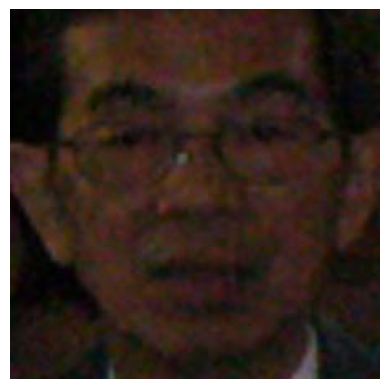}
    \centering
    \caption{Example image from \texttt{FairFace}. }
    \label{fig:fair_example}
\end{figure}

Three datasets were selected for this work to capture a diversity of image types. They were \texttt{Fashion Product}, \citep{param_aggarwal_2019}, \texttt{FairFace} \citep{karkkainen2019fairface}, and \texttt{Dollar Street} \citep{rojas2022the}. Concepts on \texttt{Fashion Product} and \texttt{Dollar Street} were selected based on nouns with large sample sizes. For \texttt{FairFace}, no sub selection of the dataset was used since they are all faces. Thus, entity extraction happened one time for all three concepts as the full dataset was covered. For all concepts, we took the first 2000 images if more than 2000 images were present. Example images and captions can be found in Figures \ref{fig:tv_example} -- \ref{fig:fair_example}.

\textbf{Figure \ref{fig:tv_example} Generated Caption:} ``The image depicts a cozy living room with a television and two chairs placed in front of it. The television is mounted on a wooden cabinet, which serves as a decorative piece in the room. The two chairs are positioned close to the television, creating a comfortable viewing area for the viewer.  In addition to the television and chairs, there is a window in the room, allowing natural light to enter''

\textbf{Extracted Entities:} `TV', `window', `cabinet', `chair'

\textbf{Figure \ref{fig:home_example} Generated Caption:} ``The image depicts a small hut with a thatched roof, situated in a dirt area. The hut is surrounded by various objects, such as pots, buckets, and other household items. There are several pots placed around the hut, some closer to the entrance and others further away. Additionally, there are two buckets positioned near the hut, one on the left side and another on the right side''

\textbf{Extracted Entities:} `pot', `household', `hut', `thatched', `entrance', `side', `area', `bucket', `items', `dirt'

\textbf{Figure \ref{fig:topwear_example} Generated Caption:} ``In the image, a young man is standing against a white background, wearing a plaid shirt and jeans.''

\textbf{Extracted Entities:} `young man', `white background', `plaid shirt', `jeans'\\

\textbf{Figure \ref{fig:jewelry_example} Generated Caption:} ``This is a gold and white diamond bracelet, featuring an intricate design that resembles a knot.''

\textbf{Extracted Entities:} `bracelet', `gold', `design', `diamond', `white', `intricate', `knot'

\textbf{Figure \ref{fig:fair_example} Generated Caption:} ``In the image, there is a man wearing glasses and a suit, standing in front of a blurry background.''

\textbf{Extracted Entities:} `man wearing', `glasses', `suit', `standing', `blurry background'

\section{Jewelry Pseudolabel Results for All Models}
\label{sec:jewelry-pseudo}
\begin{figure}[h]
  \centering
  \includegraphics[width=0.48\linewidth]{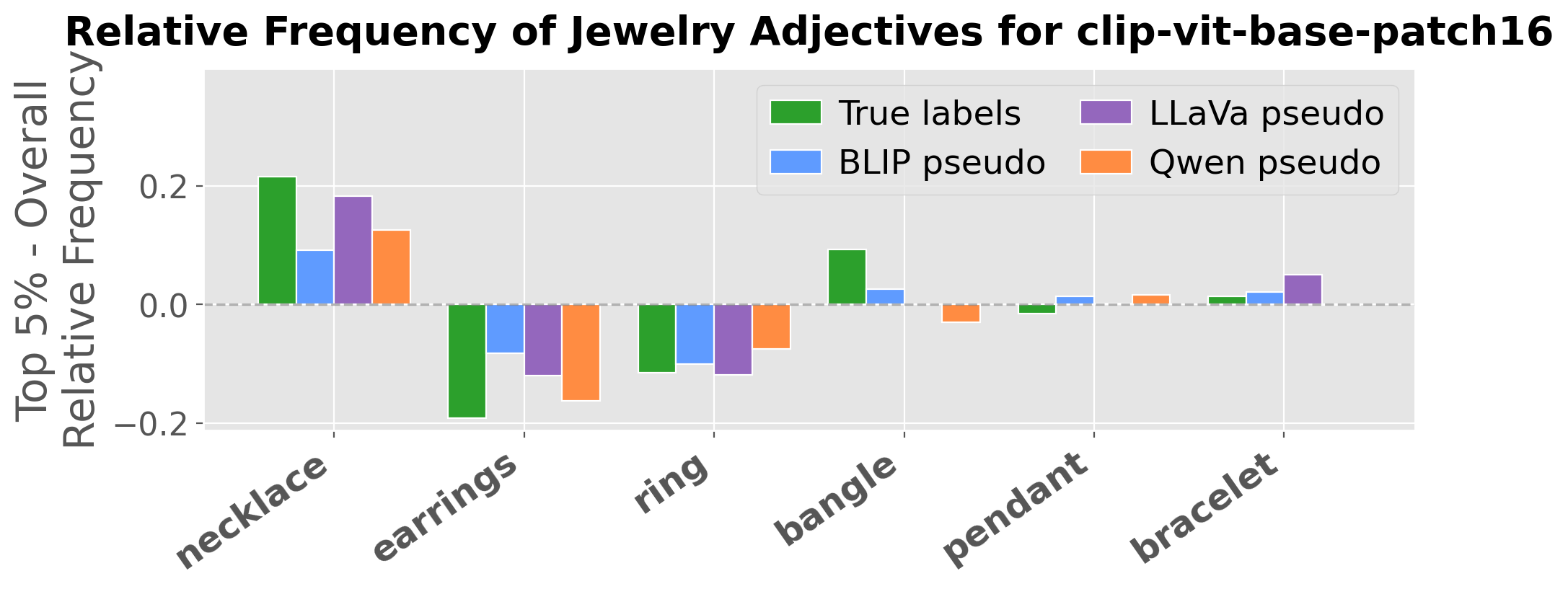}
  \hfill
  \includegraphics[width=0.48\linewidth]{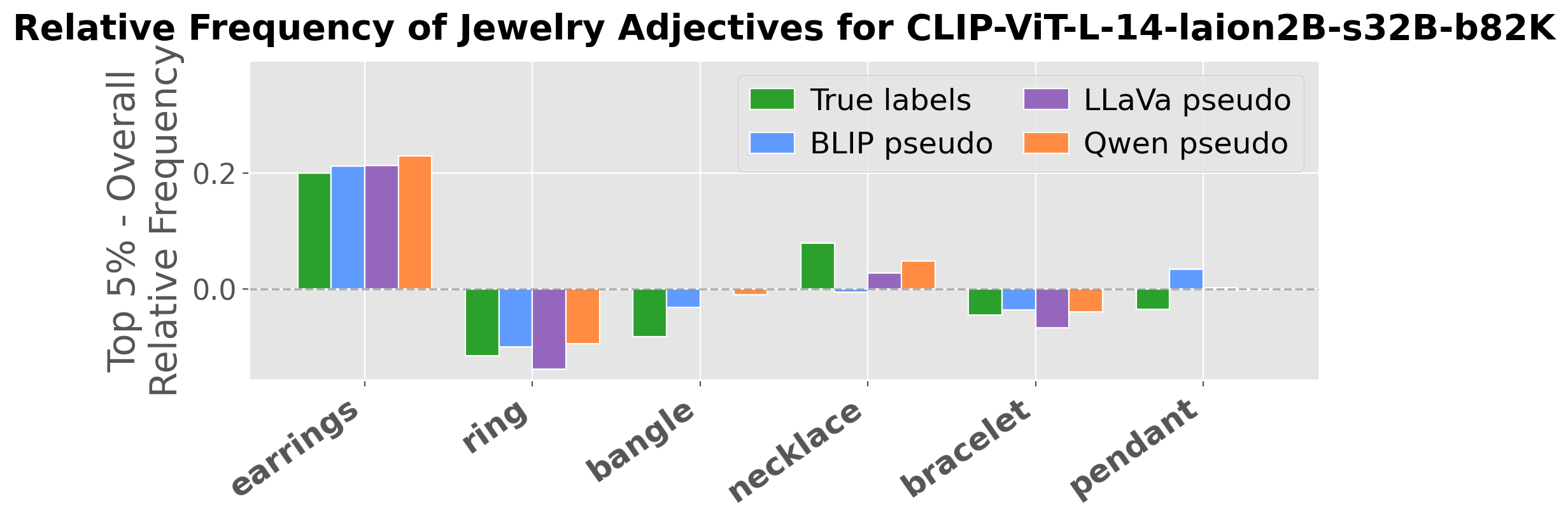}\\[0.5em]
  \includegraphics[width=0.48\linewidth]{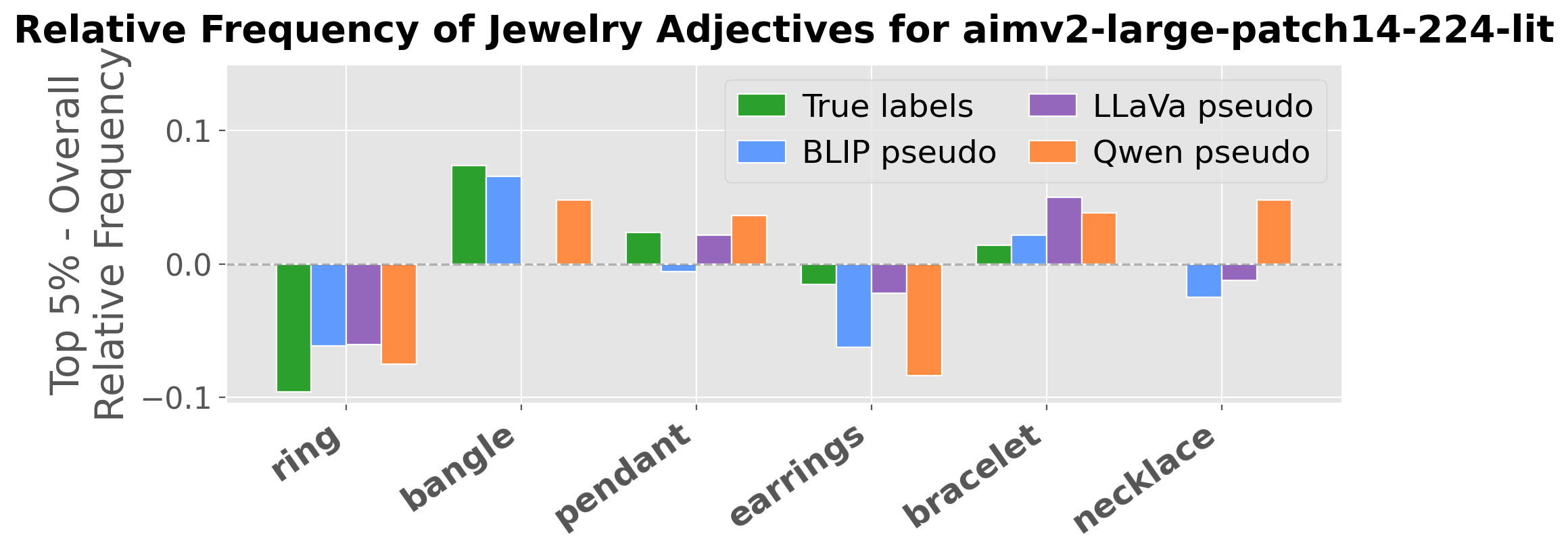}
  \hfill
  \includegraphics[width=0.48\linewidth]{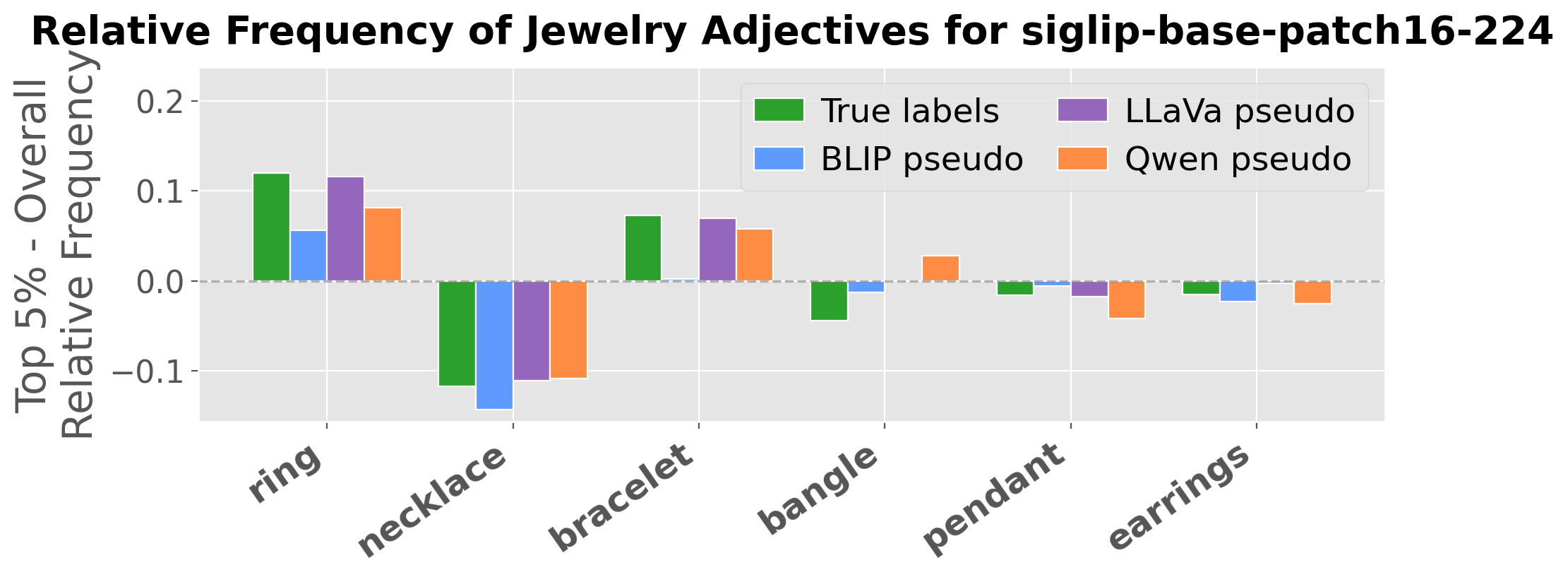}\\[0.5em]
  \caption{Pseudolabels and true label frequency for sub categories in the Jewelry concept. We find that \methodname is able to accurately identify true labels.}
  \label{fig:jewelry-sub-full}
\end{figure}

We include the full results from Figure \ref{fig:psuedo-jewelery} for all models used in this work. While individual models differ, Figure \ref{fig:jewelry-sub-full} demonstrates the similarity of true labels and pseudolabels across all models. 

\section{Results Across Models and Concepts}
\label{sec:all-concept-results}
The following Figures \ref{fig:first} -- \ref{fig:last} show frequency shift results for all datasets and queries.

\begin{figure}[h]
\centering
      \includegraphics[width=1\linewidth]{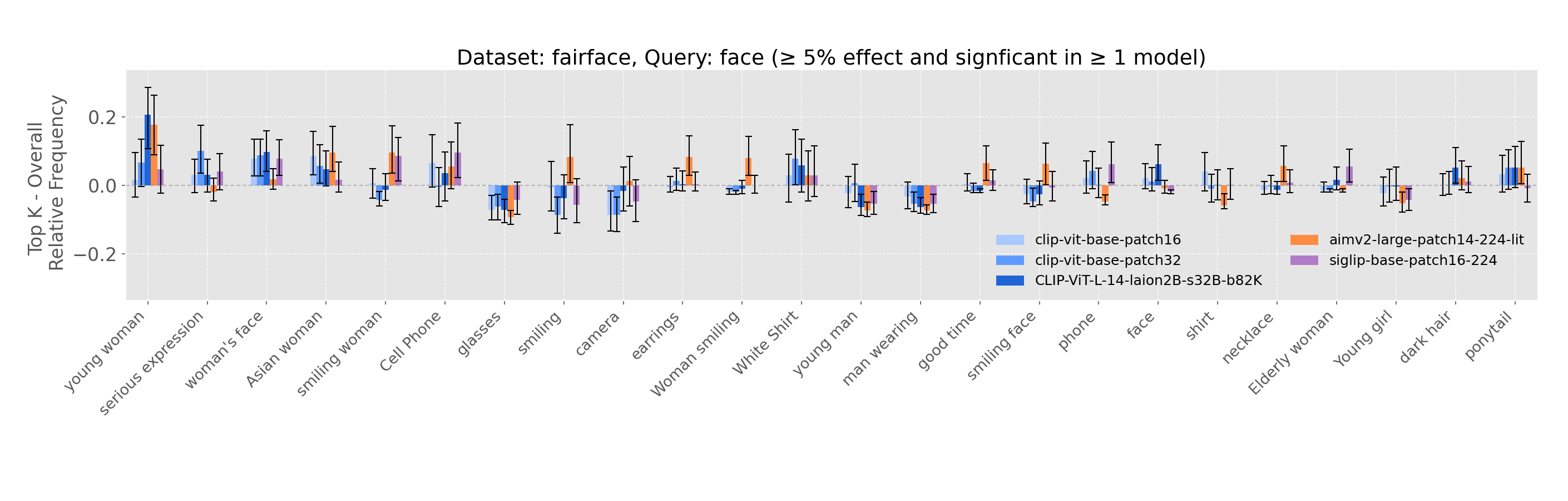}
    \caption{Frequency shift results for the `face' concept in \texttt{FairFace}. }
  \label{fig:first}
\end{figure}

\begin{figure}
    \centering
    \includegraphics[width=1\linewidth]{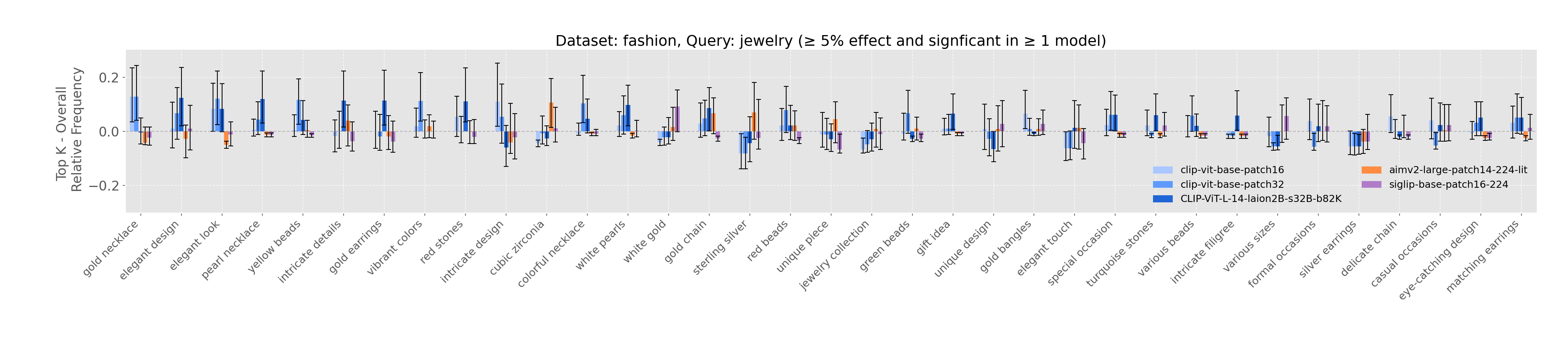}
   
  \caption{Frequency shift results for the `jewelry' concept in \texttt{Fashion Data}.}
\end{figure}

\begin{figure}[h!]
    \centering

    \includegraphics[width=1\linewidth]{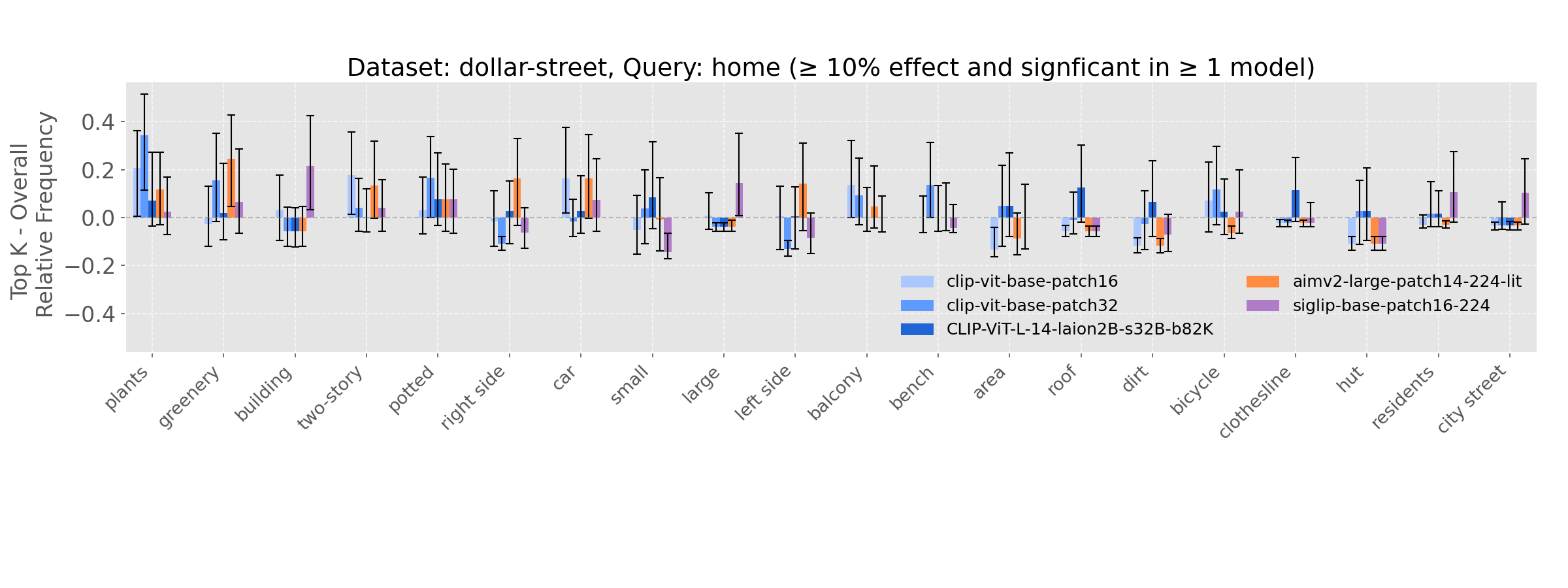}
    \caption{Frequency shift results for `home' concept in \texttt{Dollar Street}.}
\end{figure}

\begin{figure}[h!]
            \centering
            \includegraphics[width=1\linewidth]{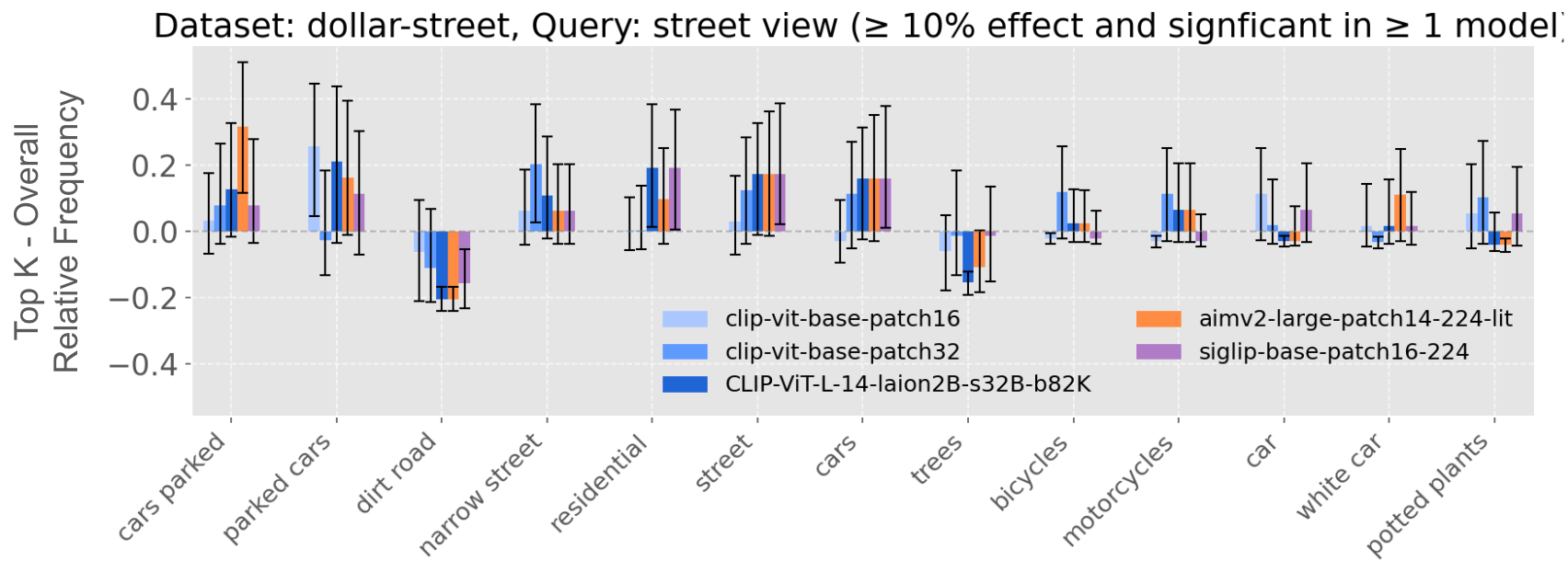}
                \caption{Frequency shift results for `street view' concept in \texttt{Dollar Street}.}
\end{figure}

\begin{figure}
       \centering
       \includegraphics[width=1\linewidth]{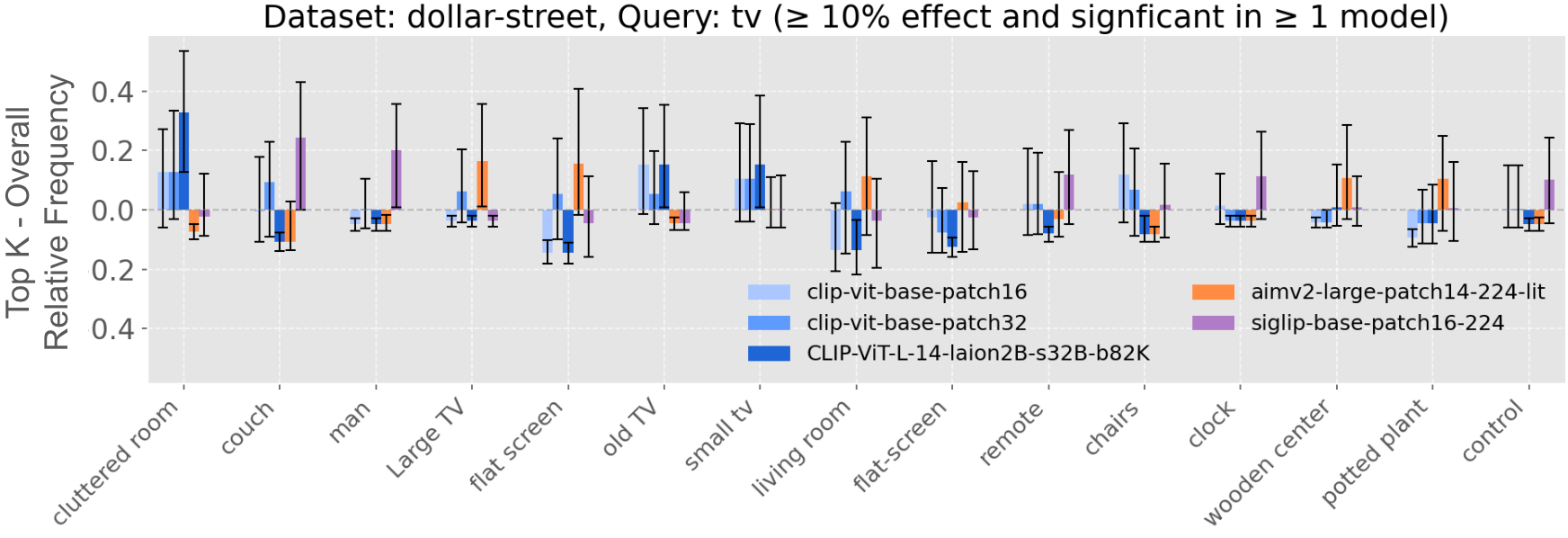}
        \caption{Frequency shift results for `TV' concept in \texttt{Dollar Street}.}
\end{figure}

\begin{figure}[h!]
  \centering
    \includegraphics[width=1\linewidth]{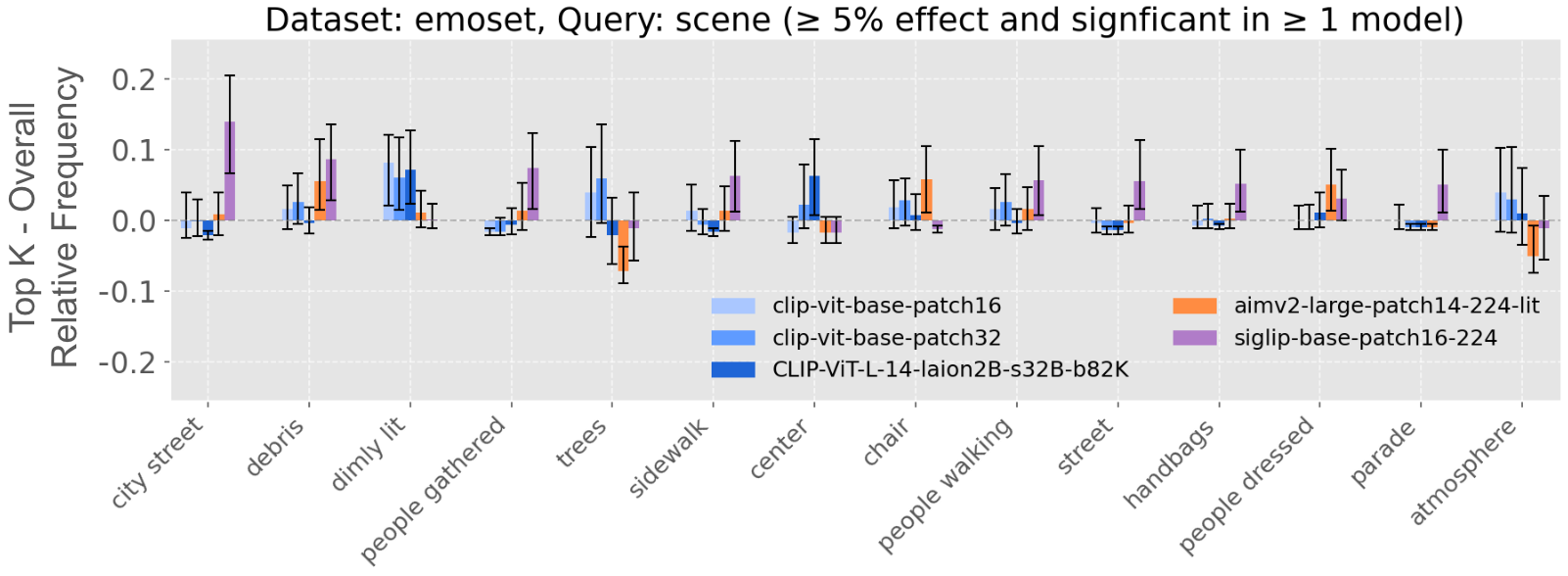}
    \caption{Frequency shift results for the `scene' concept in \texttt{EmoSet}.}
\end{figure}

\begin{figure}[h!]
  \centering
    \includegraphics[width=1\linewidth]{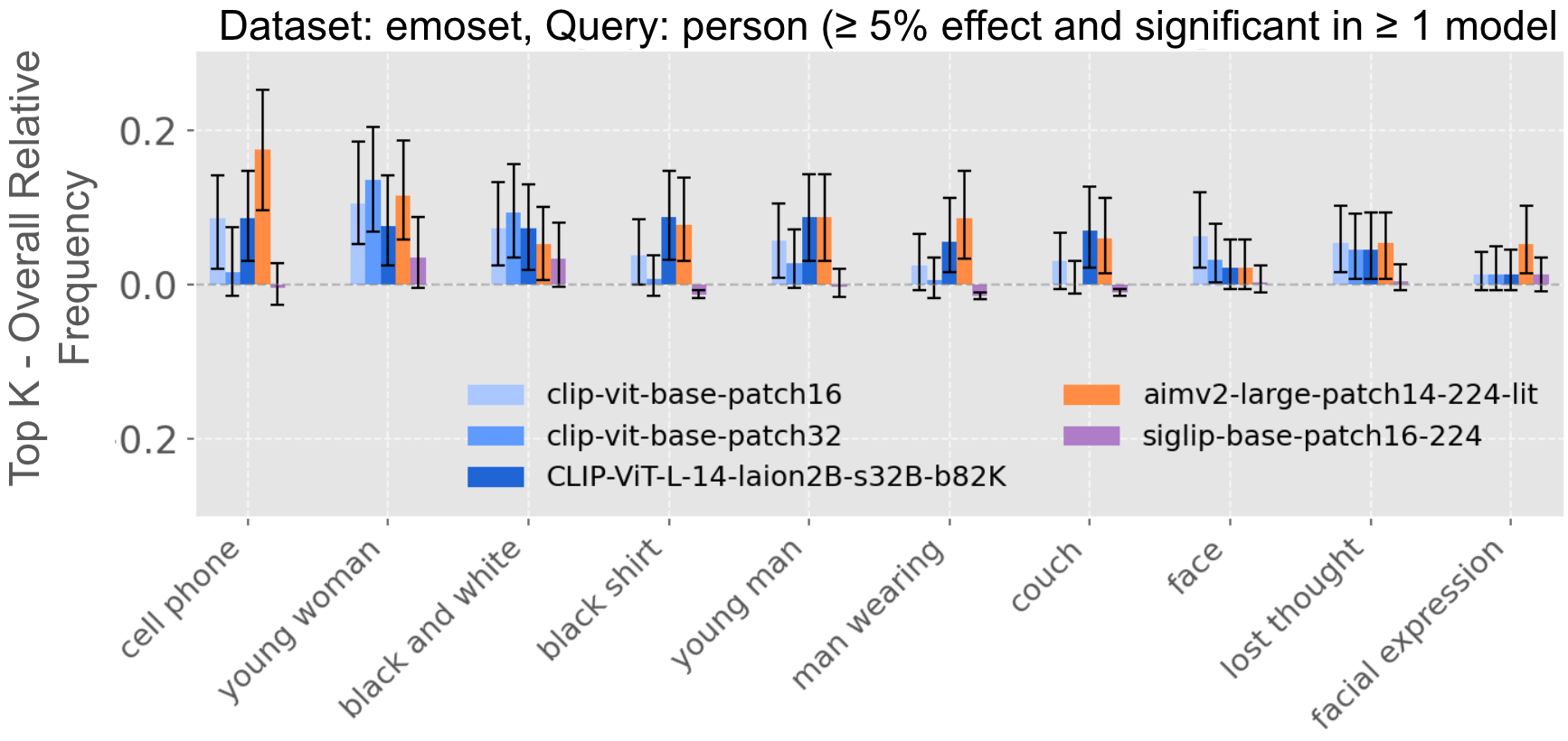}
  \caption{Frequency shift results for the `person' concept in \texttt{EmoSet}.}
  \label{fig:last}
\end{figure}

\FloatBarrier
\section{Analysis}
We formalize the analysis from the Methods section to present the mathematical guarantees of the biases identified by \methodname when using a biased caption model. Theorem \ref{thm:sign} shows that with high probability, the direction of bias is preserved even with noisy pseudolabels. We then state Corollary \ref{cor:bound_overrep} that presents a bound for how much noisy pseudolabels may overestimate the identified bias direction. This motivates the usage of \methodname as a robust spurious correlation detection framework even under imperfect setup conditions. 

\begin{theorem}[High-Probability Sign Consistency under Noisy Pseudolabels]\label{thm:sign}
Let $e$ be an entity with true overrepresentation 
\[
\Delta_{\text{true}}(e) := p_{\text{true}}^{\mathcal{D}_{\text{top-}k}}(e) - p_{\text{true}}^{\mathcal{D}}(e),
\]
and pseudolabel noise parameters $\alpha_e = \mathbb{E}[\hat{\ell}_e(x) \mid \ell_e(x) = 1]$, $\beta_e = \mathbb{E}[\hat{\ell}_e(x) \mid \ell_e(x) = 0]$.

Let $\Delta_{\text{obs}}(e)$ be the empirical observed overrepresentation computed from noisy pseudolabels over datasets $\mathcal{D}_{\text{top-}k}$ of size $k$, and $\mathcal{D}$ of size $n$.

Then with probability at least $1 - \delta$:
\[
\left| \Delta_{\text{obs}}(e) - (\alpha_e - \beta_e) \cdot \Delta_{\text{true}}(e) \right| 
\leq \epsilon, \quad 
\]
\[
\text{where } 
\epsilon := \sqrt{ \frac{1}{2} \left( \frac{1}{k} + \frac{1}{n} \right) \log\left( \frac{4}{\delta} \right) }.
\]

In particular, if
\[
|(\alpha_e - \beta_e) \cdot \Delta_{\text{true}}(e)| > \epsilon,
\]
then
\[
\operatorname{sign}(\Delta_{\text{obs}}(e)) = \operatorname{sign}(\Delta_{\text{true}}(e)),
\]
and the 
bias 
direction 
is preserved with 
probability $> 1 - \delta$.
\end{theorem}

\begin{corollary}[Bounding True Overrepresentation from Noisy Observations] 
\label{cor:bound_overrep}
Under the assumptions of Theorem \ref{thm:sign}, define the estimator:
\[
\widehat{\Delta}_{\text{true}}(e) := \frac{\Delta_{\text{obs}}(e)}{\alpha_e - \beta_e}.
\]
Then with probability at least $1 - \delta$:
\[
\left| \widehat{\Delta}_{\text{true}}(e) - \Delta_{\text{true}}(e) \right| 
\leq \frac{1}{|\alpha_e - \beta_e|} \cdot \sqrt{ \frac{1}{2} \left( \frac{1}{k} + \frac{1}{n} \right) \log \left( \frac{4}{\delta} \right) }.
\]

In particular, if $\alpha_e > \beta_e$ and $\Delta_{\text{true}}(e) > 0$, then 
$\widehat{\Delta}_{\text{true}}(e)$ provides an upwardly noisy but consistent estimate of the true frequency bias.
\end{corollary}

\section{Full Table for Competitive Analysis}
Full version of the competitive analysis Table \ref{tab:compare} is shown here in Table \ref{tab:compare-full}.

\begin{table}[h]
\centering
\caption{\label{tab:compare-full}Comparison of 
methods with \texttt{clip-base-patch-16}. \texttt{Distill} and \texttt{RaVL (Classify)}  require sub labels and thus cannot be applied to all datasets.} 
\scriptsize
\setlength{\tabcolsep}{2.5pt}
\begin{tabular}{@{}ccccccc@{}}
\toprule
                            & \# Entities & \texttt{Domino} & \texttt{RaVL} (Retrieve) & \texttt{Distill} & \texttt{RaVL} (Classify) & \methodname \\ \midrule
\multirow{4}{*}{\shortstack{Fashion,\\Jewelry}} 
                            & 2          & 0.61 [0.54, 0.68] & 0.55 [0.50, 0.58] & \textbf{0.63 [0.56, 0.71]} & 0.59 [0.54, 0.63] & 0.56 [0.48, 0.65] \\
                            & 4          & 0.68 [0.61, 0.75] & 0.67 [0.63, 0.71] & 0.67 [0.58, 0.75] & 0.66 [0.57, 0.66] & \textbf{0.70 [0.63, 0.77]}     \\
                            & 8          & 0.73 [0.65, 0.79] & 0.67 [0.63, 0.70] &  \textbf{0.78 [0.72, 0.84]}  & 0.72 [0.68, 0.75] &  0.75 [0.68, 0.81]    \\
                            & 16         & 0.73 [0.66, 0.80]  & 0.69 [0.65, 0.72] &  0.80 [0.74, 0.87]  & 0.74 [0.70, 0.77] & \textbf{0.93 [0.90, 0.97]} \\ \midrule
\multirow{4}{*}{\shortstack{EmoSet,\\Person}} 
                            & 2          & 0.77 [0.72, 0.81] & 0.50 [0.47, 0.53] &  0.76 [0.71, 0.81] & 0.60 [0.57, 0.63] &  \textbf{0.86 [0.81, 0.90]} \\
                            & 4          & 0.85 [0.82, 0.89] & 0.73 [0.71, 0.75] & 0.83 [0.79, 0.87] & 0.73 [0.71, 0.76] & \textbf{0.94 [0.91, 0.96] }    \\
                            & 8          & 0.864 [0.82, 0.90] & 0.73 [0.71, 0.76] & 0.87 [0.83, 0.91]  & 0.74 [0.71, 0.76] &  \textbf{0.95 [0.93, 0.91]}     \\
                            & 16         & 0.86 [0.82, 0.91]  & 0.75 [0.72, 0.77] & 0.88 [0.84, 0.92]  & 0.79 [0.77, 0.81] & \textbf{0.97 [0.95, 0.98]} \\ \midrule
\multirow{4}{*}{\shortstack{Dollar\\Street, TV}}
                            & 2          & 0.67 [0.55, 0.78] & 0.54 [0.48, 0.61] & - & - & \textbf{0.89 [0.80, 0.96]}     \\
                            & 4          & 0.53 [0.40, 0.64] & 0.50 [0.45, 0.57] &  -  & - & \textbf{0.87 [0.76, 0.95]}  \\
                            & 8          & 0.74 [0.62, 0.84] & 0.59 [0.53, 0.65] &  - & - & \textbf{0.92 [0.86, 0.96]}     \\
                            & 16         & 0.82 [0.73, 0.90] & 0.61 [0.55, 0.67] & - & - & \textbf{0.92 [0.84, 0.96]}     \\ \midrule
\multirow{4}{*}{\shortstack{FairFace,\\Face}} 
                            & 2          & 0.63 [0.58, 0.69] & 0.51 [0.48, 0.54] & - & - & \textbf{0.83 [0.78, 0.87]}     \\
                            & 4          & 0.71 [0.65, 0.76] & 0.57 [0.55, 0.60] & -  & - & \textbf{0.88 [0.84, 0.91]}  \\
                            & 8          &  0.73 [0.67, 0.78] & 0.66 [0.63, 0.69] &  - & - & \textbf{0.90 [0.87, 0.94]} \\
                            & 16         & 0.72 [0.67, 0.77] & 0.67 [0.64, 0.70] &  - & - & \textbf{0.97 [0.96, 0.98]}  \\ \midrule
\multirow{4}{*}{\shortstack{Dollar\\Street, Home}} & 2          & 0.59 [0.46, 0.72] & 0.53 [0.47, 0.60] & - & - & \textbf{0.80 [0.72, 0.88]}     \\
                            & 4          & 0.72 [0.58, 0.84] & 0.52 [0.46, 0.59] &  -  & - & \textbf{0.81 [0.70, 0.92]}  \\
                            & 8          & 0.75 [0.64, 0.86] & 0.56 [0.51, 0.63] &  - & - & \textbf{0.84 [0.73, 0.94]}     \\
                            & 16         & 0.79 [0.70, 0.87] & 0.56 [0.50, 0.62] & - & - & \textbf{0.87 [0.79, 0.94]}     \\ \midrule
\multirow{4}{*}{\shortstack{Dollar\\Street, Street View}} 
                            & 2          & 0.67 [0.54, 0.79] & 0.53 [0.47, 0.59] & - & - & \textbf{0.75 [0.65, 0.83]}     \\
                            & 4          & 0.78 [0.68, 0.85] & 0.65 [0.59, 0.71] &  -  & - & \textbf{0.88 [0.81, 0.93]}  \\
                            & 8          & 0.81 [0.72, 0.88] & 0.69 [0.62, 0.75] &  - & - & \textbf{0.91 [0.83, 0.97]}     \\
                            & 16         & 0.84 [0.76, 0.91] & 0.77 [0.71, 0.82] & - & - & \textbf{0.92 [0.85, 0.97]}     \\ \midrule
\end{tabular}
\end{table}

\section{Inserted Bias Recovery}
\begin{figure}[t]
  \centering
  \begin{subfigure}[t]{0.45\linewidth}
    \centering
    \includegraphics[width=\linewidth]{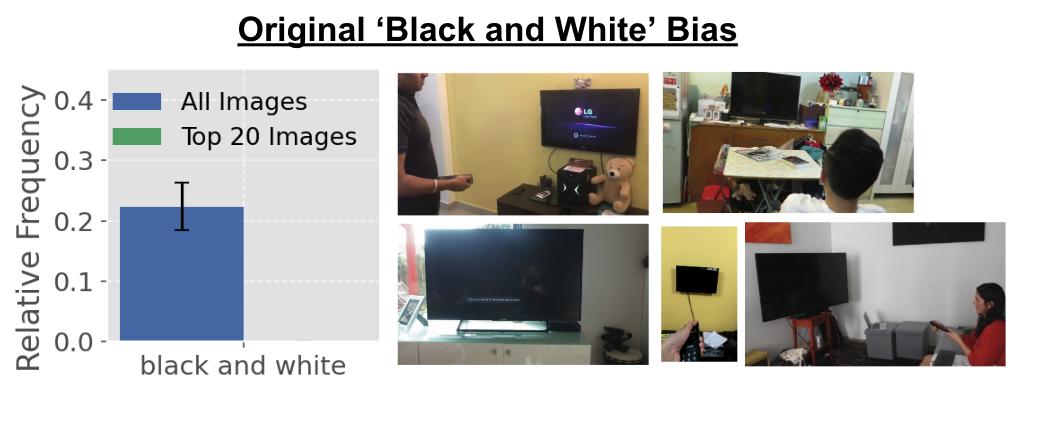}
    \caption{Original bias distribution of black and white entity.}
    \label{fig:bw-before}
  \end{subfigure}
  \hfill
  \begin{subfigure}[t]{0.45\linewidth}
    \centering
    \includegraphics[width=\linewidth]{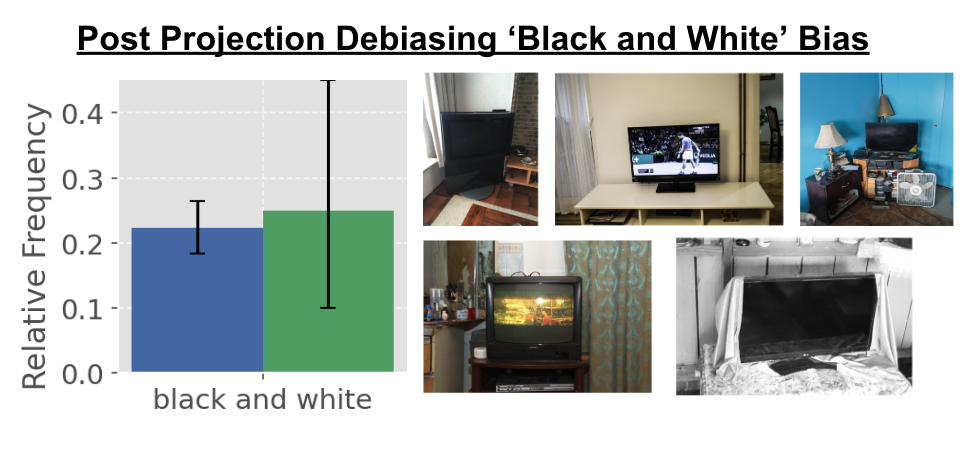}
    \caption{Post-projection debiasing black and white entity distribution.}
    \label{fig:bw-after}
  \end{subfigure}
  \caption{\textbf{\methodname identifies grayscale bias and enables its correction.
  } On \texttt{Dollar Street} TVs with 50\% of images randomly converted to grayscale, querying \texttt{clip-vit-base-patch-16} for ``expensive'' underrepresents 
  grayscale
  images in the top-20 retrieved (a); projection debiasing on the black-and-white concept restores the base-rate distribution (b).}
  \label{fig:black-and-white}
\end{figure}

We inserted a bias on Dollar Street's TV images by randomly perturbing 50\% of images to grayscale, ran the full \methodname pipeline, and queried for ``expensive'' on \texttt{clip-vit-base-patch-16}. \methodname correctly identified 
``black-and-white'' as a relevant element, and determined that this was the most biased feature:
in the top-20 retrieved images
none were black-and-white despite half of the images in the dataset being black-and-white (see Figure \ref{fig:black-and-white}).  

We demonstrate that the model was indeed relying on the presence or absence of grayscale when determining rankings, 
by using projection-based debiasing \citep{chuang2023debiasing, bolukbasi2016man} to make the query embedding orthogonal to the black-and-white concept. After projection debiasing, the model had approximately equal relative frequency for the target concept Figure \ref{fig:black-and-white}. This analysis shows that \textbf{\methodname captures visual attributes and map those attributes to corresponding biases in VLMs}.

\section{Compute}
All experiments were run on a GPU on an internal cluster. While multiple GPUs may have been used to speed up the slower pieces of experimentation, mainly captioning and entity extraction, multiple GPUs not necessary for the implementation of \methodname.

\end{document}